\documentclass[11pt]{article}

\usepackage[preprint]{acl}

\usepackage{times}
\usepackage{latexsym}

\usepackage[T1]{fontenc}

\usepackage[utf8]{inputenc}

\usepackage{microtype}
\usepackage{enumitem}

\usepackage{inconsolata}

\usepackage{graphicx}
\usepackage{color}
\usepackage{algorithm}
\usepackage{algorithmic}
\usepackage{booktabs} 
\usepackage{makecell}
\usepackage{multirow}

\usepackage{amsmath}
\usepackage{amsfonts}
\usepackage{amssymb}

\title{Memory-Augmented LLM-based Multi-Agent System for Automated Feature Generation on Tabular Data}

\author{
  \textbf{Fengxian Dong\textsuperscript{1}},
  \textbf{Zhi Zheng\textsuperscript{1}\thanks{Corresponding authors.}},
  \textbf{Xiao Han\textsuperscript{2}},
  \textbf{Wei Chen\textsuperscript{1}}, \\
  \textbf{Jingqing Ruan\textsuperscript{3}},
  \textbf{Tong Xu\textsuperscript{1}\footnotemark[1]},
  \textbf{Yong Chen\textsuperscript{1}},
  \textbf{Enhong Chen\textsuperscript{1}} \\
\textsuperscript{1}University of Science and Technology of China \\
\textsuperscript{2}Zhejiang University of Technology,
\textsuperscript{3}Meituan \\
\{fengxiandong, chenweicw, chenyong1997\}@mail.ustc.edu.cn \\
\{zhengzhi97, tongxu, cheneh\}@ustc.edu.cn\\
hahahenha@gmail.com, ruanjingqing2019@ia.ac.cn 
}

\begin{document}
\maketitle

\begin{abstract}
Automated feature generation extracts informative features from raw tabular data without manual intervention and is crucial for accurate, generalizable machine learning. 
Traditional methods rely on predefined operator libraries and cannot leverage task semantics, limiting their ability to produce diverse, high-value features for complex tasks. Recent Large Language Model (LLM)-based approaches introduce richer semantic signals, but still suffer from a restricted feature space due to fixed generation patterns and from the absence of feedback from the learning objective. 
To address these challenges, we propose a Memory-Augmented LLM-based Multi-Agent System (\textbf{MALMAS}) for automated feature generation. MALMAS decomposes the generation process into agents with distinct responsibilities, and a Router Agent activates an appropriate subset of agents per iteration, further broadening exploration of the feature space. We further integrate a memory module comprising procedural memory, feedback memory, and conceptual memory, enabling iterative refinement that adaptively guides subsequent feature generation and improves feature quality and diversity. Extensive experiments on multiple public datasets against state-of-the-art baselines demonstrate the effectiveness of our approach. The code is available at \url{https://github.com/fxdong24/MALMAS}
\end{abstract}

\begin{figure}[!t]
\centering
\includegraphics[width=1\columnwidth]{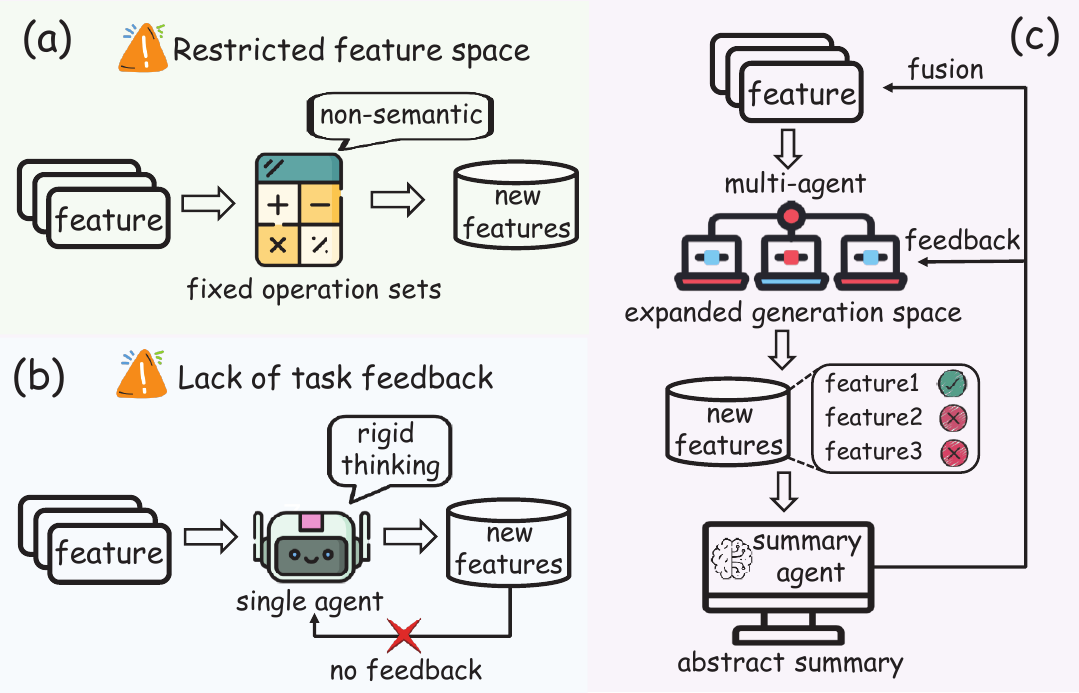} 
\caption{
Comparison of traditional, LLM-based, and multi-agent feature generation approaches.
}
\vspace{-3mm}

\label{Motivating}
\end{figure}

\section{Introduction}

Recently, the advancement of Automated Machine Learning (\textbf{AutoML}) has greatly improved the efficiency of data modeling~\cite{auto:1,auto:2,auto:3,auto:4,auto:5}. Within this paradigm, \emph{automated feature generation}, which extracts informative features from raw data without manual intervention, has become a key enabler for building accurate and generalizable models.

However, traditional automated feature generation methods still suffer from several limitations, which hinder their ability to produce high-quality features effectively.
As illustrated in Figure~\ref{Motivating}(a), these methods apply a predefined set of operators to original features to construct new feature sets~\cite{baseline:2,baseline:3,baseline:4}.
They rely on limited operator sets and do not incorporate feature semantics, which confines the transformation search space to a narrow region.

Recently, large language models (LLMs) have shown strong semantic understanding and generation capabilities~\cite{auto:9}, motivating LLM-based feature generation that leverages task descriptions to propose transformations~\cite{baseline:1,auto:6}. However, these methods typically rely on static, singular generation strategies rooted in rigid thinking, which still constrain the exploration of broader feature spaces.
More critically, these methods lack mechanisms to adapt generation strategies based on historical experience or task-specific feedback. Without such adaptive signals, feature generation becomes disconnected from learning performance, leading to inefficient, trial-and-error exploration and limited ability to prioritize high-value transformations.
This feedback-insensitive process undermines learning-goal alignment and hampers methods’ performance, as illustrated in Figure~\ref{Motivating}(b).

To this end, we propose the Memory-Augmented LLM-based Multi-Agent System (\textbf{MALMAS}), an automated feature generation framework, as illustrated in Figure~\ref{Motivating}(c).

Specifically, we decompose feature generation into multiple independent agents with roles grounded in a principled framework along three largely orthogonal dimensions from established feature engineering practice: transformation complexity, data scope, and data-type dependency. Inspired by the categorization of high-value ``golden features''~\citep{baseline:5}, this design assigns each agent a clear, specialized responsibility. A Router Agent dynamically selects a subset of agents from a predefined pool based on task metadata and accumulated memory, enabling adaptive allocation of generation effort. Each agent then conducts multi-turn interactions with the LLM, constructing role-specific prompts conditioned on the current feature set and experiential feedback. By exploring complementary regions of the feature space, the agents mitigate feature homogenization in static single-agent strategies~\cite{baseline:1} and reduce functional redundancy, thereby broadening the overall search space.

To address the lack of feedback-driven adaptability, we equip MALMAS with a multi-level memory that enables credit assignment and strategy updates across rounds. Procedural memory caches executed transformations to suppress redundant exploration, feedback memory attributes validation utility to generated features, and conceptual memory abstracts reusable heuristics from historical traces for longer-horizon adaptation. A Summary Agent aggregates cross-agent feedback and concepts into a global conceptual memory that conditions subsequent routing and prompting. This design turns per-round evaluations into persistent learning signals, steering generation toward high-yield, task-relevant transformations.

The main contributions of this paper are summarized as follows:
\begin{itemize}[topsep=2pt,itemsep=1pt,parsep=0pt,partopsep=0pt,leftmargin=*]
    \item We propose the first multi-agent framework for automated feature generation, enabling collaborative exploration beyond predefined operators and improving feature diversity.

    \item We develop a multi-level memory mechanism that integrates procedural traces, feedback, and conceptual abstractions, allowing agents to iteratively refine their strategies.
    \item We evaluate MALMAS on 16 classification and 7 regression datasets, where it outperforms baselines, and we provide practical analyses demonstrating real-world applicability and efficiency. 
\end{itemize}

\section{Related Work}

\subsection{Traditional and LLM-enhanced AutoML}


Before the emergence of LLMs, end-to-end AutoML systems such as Auto-WEKA, Auto-sklearn, H2O AutoML, Google AutoML Tables, and FLAML~\cite{nollm:1,nollm:2,nollm:3,nollm:4,nollm:5,auto:10} mainly focused on pipeline search, hyperparameter optimization, and model selection.

With the advent of LLMs, AutoML has increasingly adopted natural language as an interface for automation. Methods such as Text-to-ML, LLM-Select, DS-Agent, and GL-Agent~\cite{auto:1,auto:2,auto:3,auto:4,auto:5} employ LLMs to generate or recommend ML pipelines, leveraging instruction following and agentic workflows to reduce manual pipeline design.

However, most of these systems emphasize model and pipeline configuration, while feature construction remains limited to basic preprocessing operations~\cite{auto:7}. This indicates a gap between LLM-enhanced AutoML pipelines and domain-aware feature engineering.

\subsection{Automated Feature Generation}

Feature generation is a long-standing and critical step for improving model performance. Traditional methods such as autofeat~\cite{baseline:2}, Deep Feature Synthesis (DFS)~\cite{baseline:3}, and OpenFE~\cite{baseline:4} apply symbolic transformations over predefined operator sets. DFS, implemented in Featuretools, demonstrates the practicality of compositional operators by automatically generating features from relational data. While these methods are efficient and interpretable, they are constrained by fixed operator libraries and limited adaptation to task-specific semantics.
More recently, LLM-based methods enable semantically driven generation~\cite{baseline:1,auto:6,baseline:6}. CAAFE~\cite{baseline:1} uses task descriptions to better align generated features with downstream objectives, and OCTree combines LLMs with tree-based reasoning to support feature validation and interpretability.



\subsection{Multi-Agent Systems}
LLM-based multi-agent systems have emerged as a promising paradigm for collaboration, specialization, and iterative reasoning~\cite{app:9,app:10}. They have been applied to social simulation (e.g., Generative Agents and AgentSociety~\cite{app:1,app:15}), software development (e.g., AutoGen and CodeAct~\cite{app:8,app:6}), and decision-making via multi-agent debate with sparse communication~\cite{app:12,app:13,app:14,li2026dynadebatebreakinghomogeneitymultiagent}. Recent methods such as ReAct and Reflexion further highlight the role of memory and feedback-driven reasoning-action loops for continual improvement~\cite{app:16,app:17}. Moreover, memory has been shown to be crucial for retaining useful experience and improving long-horizon decision-making~\cite{liu,xu2026from}.
Despite these advances, multi-agent systems for automated feature generation remain underexplored.


\begin{figure*}[t]
\centering
\includegraphics[width=0.95\textwidth]{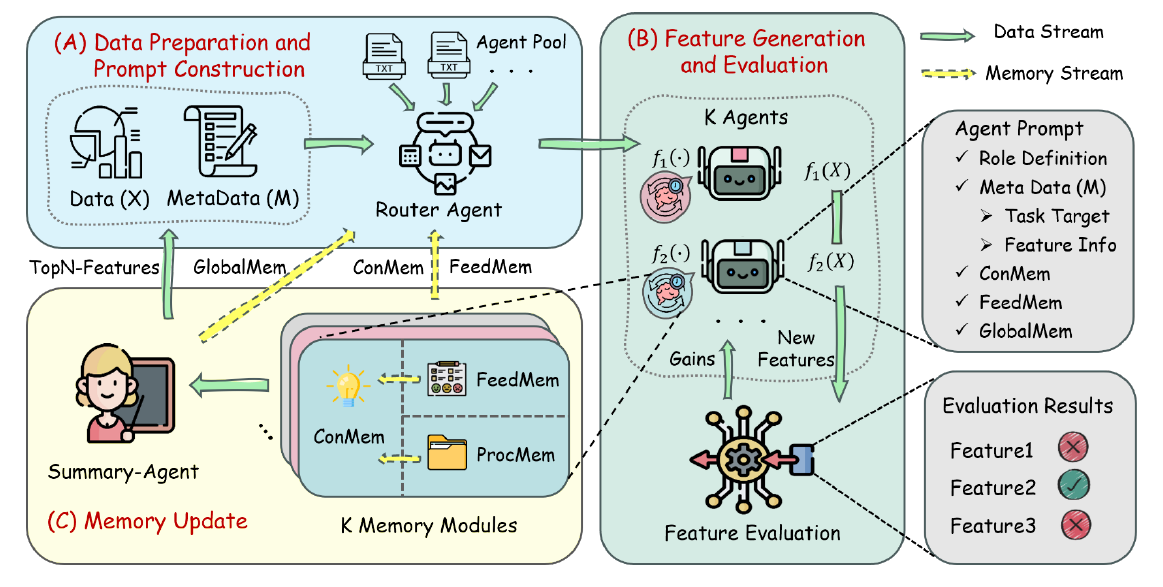}
\caption{
Overview of the proposed MALMAS framework.
(A) Agents construct prompts from metadata, selected features, and memory.
(B) Agents generate and evaluate candidate features with a downstream model.
(C) Agent memories are summarized into a global memory to guide subsequent rounds.
}
\vspace{-2mm}

\label{figure:pipeline}
\end{figure*}

\section{Problem Formulation}

Given a labeled tabular dataset $\mathbb{D} = (X, y)$, where $X \in \mathbb{R}^{m \times n}$ is the feature matrix with $m$ instances and $n$ features, and $y \in \mathbb{R}^m$ is the corresponding label vector. The goal of feature generation is to find a transformation function $T: X \to \widetilde{X}$, where $\widetilde{X} = X \cup T(X)$, that improves the predictive performance of a model $f$ when trained on the enhanced feature space. Formally, the objective is to maximize the validation performance of $\mathcal{F}$:
{\setlength{\abovedisplayskip}{4pt}
 \setlength{\belowdisplayskip}{4pt}
 \setlength{\abovedisplayshortskip}{4pt}
 \setlength{\belowdisplayshortskip}{4pt}
\begin{equation}
T^* = \arg \max_{T} \mathcal{E}(\mathcal{F}( X_{\text{val}} \cup T(X_{\text{val}})), Y_{\text{val}}),
\end{equation}}
where $\mathcal{E}$ is the evaluation metric, and $(X_{\text{val}}, Y_{\text{val}})$ is the validation set from cross-validation. Here, $T^*$ denotes the optimal transformation function that produces feature beneficial for model performance.


\section{Methodology}

Feature engineering for tabular data requires diverse, context-aware transformations, yet most automated methods rely on a single strategy or weakly coupled modules, limiting broad, task-relevant exploration. We propose \textbf{MALMAS}, which coordinates specialized agents to generate and refine features. With role-specific agents and shared procedural, feedback, and conceptual memories, MALMAS enables iterative exploration and underpins the pipeline in Figure~\ref{figure:pipeline}.

\subsection{Multi-agent Structure}

To address the limited ability of a single generator to deeply explore novel features, MALMAS maintains a pool of specialized agents and employs a Router Agent to activate an appropriate subset per iteration. This design increases the diversity and adaptability of generated features while avoiding unnecessary exploration by inapplicable strategies.

\subsubsection{Parallel Generation Architecture}
MALMAS maintains an agent pool $\mathcal{A}=\{A_i\}_{i=1}^{K}$, where each agent implements a distinct feature transformation strategy. At iteration $r$, a Router Agent selects an active subset $\mathcal{A}^{(r)} \subseteq \mathcal{A}$, and only the selected agents run in parallel to explore complementary feature interactions, transformations, and compositions. Over multiple rounds, this design adapts to diverse feature types and modeling needs through heterogeneous strategies. The overall process is formulated as:
{\setlength{\abovedisplayskip}{4pt}
 \setlength{\belowdisplayskip}{4pt}
 \setlength{\abovedisplayshortskip}{4pt}
 \setlength{\belowdisplayshortskip}{4pt}
\begin{equation}
    T^{(r)} = \bigcup_{A_i \in \mathcal{A}^{(r)}} T^{(r)}_i,
    \quad k_r = \left|\mathcal{A}^{(r)}\right|.
\end{equation}}
Here, $T^{(r)}$ denotes the aggregated set of features generated in the $r$-th round, and $T^{(r)}_i$ represents the subset produced by agent $A_i$ when activated.

In each round, each active agent $A_i \in \mathcal{A}^{(r)}$ independently generates a subset of new features $T^{(r)}_i$ from the current dataset $X$ by applying its designated strategy. Taking the union of the activated agents' outputs yields an enriched and more comprehensive feature space for model training.

\subsubsection{Agent Responsibilities}\label{agent:responsibilities}

To systematically explore the vast feature space, MALMAS adopts a principled multi-agent framework that decomposes feature generation along three largely orthogonal dimensions from feature engineering practice: transformation complexity, data scope, and data-type dependency. Inspired by the categorization of high-value ``golden features''~\citep{baseline:5}, this design encourages agents to explore complementary aspects of the data, increasing feature diversity while reducing functional redundancy.

Each agent $A_i$ applies a distinct transformation strategy $f_i(\cdot)$ to the dataset $X$, producing a feature subset $T_i$ aligned with its objective:
{\setlength{\abovedisplayskip}{4pt}
 \setlength{\belowdisplayskip}{4pt}
 \setlength{\abovedisplayshortskip}{4pt}
 \setlength{\belowdisplayshortskip}{4pt}
\begin{equation}
    T_i = f_i(X).
\end{equation}}
We instantiate a fixed pool of strategy agents as follows, from which the Router Agent activates a subset at each iteration:
\begin{itemize}[topsep=2pt,itemsep=1pt,parsep=0pt,partopsep=0pt,leftmargin=*]
    \item \textbf{Unary-Feature Agent}. Applies unary transformations $f_{\text{unary}}(X)$ to individual features to generate basic but informative variants.
    \item \textbf{Cross-Compositional Agent}. Combines multiple inputs $f_{\text{compositional}}(X)$ to capture higher-order interactions.
    \item \textbf{Temporal-Feature Agent}. Extracts temporal patterns $f_{\text{temporal}}(X)$ for time-series data.
    \item \textbf{Aggregation-Construct Agent}. Generates group-level summary features $f_{\text{aggregation}}(X)$.
    \item \textbf{Local-Transform Agent}. Applies region-specific transformations $f_{\text{local-transform}}(X)$ to capture locally informative patterns.
    \item \textbf{Local-Pattern Agent}. Discovers latent patterns within feature subsets via clustering or local interaction modeling $f_{\text{local-pattern}}(X)$.
\end{itemize}

\subsection{Memory Architecture and Management}
Feature generation in MALMAS is formulated as an iterative search over transformations, where learning signals from downstream evaluation are \emph{persisted} and \emph{reused} to refine future generation. As illustrated in Figure~\ref{figure:pipeline}(C), each agent maintains a structured memory state that supports cross-round \emph{credit assignment} and \emph{strategy refinement}, thereby turning expensive feedback into reusable guidance.

Formally, at iteration $r$, each agent $A_i$ maintains an explicit, structured memory state
$\mathcal{M}_i^{(r)}=\{\text{ProcMem}_i^{(r)},\text{FeedMem}_i^{(r)},\text{ConMem}_i^{(r)}\}$.
At the beginning of round $r$, the agent retrieves its local memories together with the shared $\text{GlobalMem}^{(r-1)}$ to condition prompt construction; after feature evaluation, it appends new traces and utilities to update $\mathcal{M}_i^{(r)}$. Intuitively, procedural memory captures \emph{what was tried}, feedback memory captures \emph{what worked}, and conceptual memory captures \emph{why it worked} in a compact form.

\subsubsection{Procedural Memory}
Procedural memory serves as an \emph{execution trace} that records the concrete transformation actions performed by agent $A_i$, enabling reproducibility and constraining redundant exploration.
In iteration $r$, after generating $n_i^{(r)}$ features:
{\setlength{\abovedisplayskip}{4pt}
 \setlength{\belowdisplayskip}{4pt}
 \setlength{\abovedisplayshortskip}{4pt}
 \setlength{\belowdisplayshortskip}{4pt}
\begin{equation}
    \text{ProcMem}_i^{(r)} = \left\{ \left( b_j, t_j, f_j, d_j, r \right) \right\}_{j=1}^{n_i^{(r)}},
\end{equation}}
where $b_j$ denotes the base columns, $t_j$ the transformation type, $f_j$ the generated feature name, $d_j$ the transformation description, and $r$ the iteration index. During subsequent rounds, $\text{ProcMem}_i$ is used to avoid duplicate transformations and to discourage patterns that repeatedly fail under evaluation.
\subsubsection{Feedback Memory}
Feedback memory provides a \emph{utility signal} by associating each generated feature with its downstream validation outcome, enabling explicit credit assignment for feature transformations. For agent $A_i$ in iteration $r$ with $n_i^{(r)}$ generated features:
{\setlength{\abovedisplayskip}{4pt}
 \setlength{\belowdisplayskip}{4pt}
 \setlength{\abovedisplayshortskip}{4pt}
 \setlength{\belowdisplayshortskip}{4pt}
\begin{equation}
    \text{FeedMem}_i^{(r)} = \left\{ \left( f_j, m, v_j, e_j, r \right) \right\}_{j=1}^{n_i^{(r)}},
\end{equation}}
where $f_j$ is the feature name, $m$ is the evaluation metric, $v_j$ is the metric value, $e_j$ indicates whether the feature is effective, and $r$ is the iteration index. This memory enables \emph{utility attribution} by linking each feature to validation gain, which biases later rounds toward high-yield transformations and away from noisy or low-impact candidates.
\vspace{-2mm}
\subsubsection{Conceptual Memory}
Conceptual memory stores a compact set of reusable heuristics distilled from an agent's historical traces and utilities. After each round, the LLM summarizes $\text{ProcMem}_i^{(r)}$ and $\text{FeedMem}_i^{(r)}$ into rules that guide subsequent generation:

{\setlength{\abovedisplayskip}{4pt}
 \setlength{\belowdisplayskip}{4pt}
 \setlength{\abovedisplayshortskip}{4pt}
 \setlength{\belowdisplayshortskip}{4pt}
 \fontsize{9.5pt}{10pt}\selectfont
\begin{equation}
    \text{ConMem}_i^{(r)} = \text{LLM}\left(\text{ProcMem}_i^{(r)},\text{FeedMem}_i^{(r)}\right).
\end{equation}}
By compressing experience into high-level guidance, $\text{ConMem}_i$ supports strategy adaptation across rounds keeping the prompt context concise.

\begin{algorithm}[!t]
\caption{Iterative Feature Generation}
\label{alg:malmas}
\footnotesize
\setlength{\algorithmicindent}{1.1em}

\textbf{Input}: $X^{(1)}$, $y$, metadata $M$, rounds $R$,\\ agent pool $\mathcal{A}=\{A_i\}_{i=1}^{K}$, metric $E$\\
\textbf{Output}: $X^{(R+1)}$

\begin{algorithmic}[1]
\STATE Initialize memories for all agents $A_i \in \mathcal{A}$
\FOR{$r = 1,2,\dots,R$}
    \STATE $\mathcal{A}^{(r)} \leftarrow \texttt{Route}(M, \text{GlobalMem}^{(r-1)})$
    \FOR{each active agent $A_i \in \mathcal{A}^{(r)}$}
\STATE $p_{\text{mem}} \leftarrow \{ \text{FeedMem}_i^{(r-1)}, \text{ConMem}_i^{(r-1)},$
\STATE   \hspace{\algorithmicindent}$\text{GlobalMem}^{(r-1)} \}$

        \STATE $p \leftarrow \texttt{ConstructPrompt}(M, p_{\text{mem}})$
        \STATE $T_i^{(r)} \leftarrow \pi_{\theta}(p, X^{(r)})$ \COMMENT{Generate $f_i^{(r)}$}
        \STATE $\text{gains} \leftarrow \mathcal{E}((T_i^{(r)} \cup X^{(r)}), y)$
        \STATE \texttt{Update} $(\text{ProcMem}_i^{(r)},\text{FeedMem}_i^{(r)}, \text{ConMem}_i^{(r)})$
    \ENDFOR
    \STATE $\text{Mem}^{(r)} \leftarrow \bigcup_{A_i \in \mathcal{A}^{(r)}}\!\left(\text{ConMem}_i^{(r)} \cup \text{FeedMem}_i^{(r)}\right)$
    \STATE $\text{GlobalMem}^{(r)} \leftarrow \texttt{Summary}(\text{Mem}^{(r)})$
    \STATE $T^{(r)} \leftarrow \bigcup_{A_i \in \mathcal{A}^{(r)}} T_i^{(r)}$
    \STATE $F^{(r)} \leftarrow \bigcup_{A_i \in \mathcal{A}^{(r)}} \text{FeedMem}_i^{(r)}$
    \STATE $S^{(r)} \gets \texttt{TopN-Features}(T^{(r)}, F^{(r)})$
    \STATE $X^{(r+1)} \gets X^{(r)} \cup S^{(r)}$
    \STATE $M \gets \texttt{UpdateMetadata}(M, S^{(r)})$
\ENDFOR
\RETURN $X^{(R+1)}$

\end{algorithmic}
\end{algorithm}

\subsubsection{Global Conceptual Memory}
To promote coordination and knowledge transfer across agents, after each iteration the Summary-Agent aggregates agents' local conceptual and feedback memories into a Global Conceptual Memory. This cross-agent consolidation forms a shared prior for the next round, propagating effective transformation heuristics across roles, reducing overlap among agents, and improving the efficiency of subsequent exploration and refinement.

\subsection{Iterative Feature Generation}\label{sec:3.4}

This section describes the feature generation mechanism of our multi-agent system, as shown in Figure~\ref{figure:pipeline}. Across iterative rounds, agents leverage local and global memories to refine their strategies.

In each iteration $r$, each active agent $A_i \in \mathcal{A}^{(r)}$ independently executes a fixed sequence of steps, as detailed in Algorithm~\ref{alg:malmas}:
\begin{itemize}[topsep=2pt,itemsep=1pt,parsep=0pt,partopsep=0pt,leftmargin=*]

\item \textbf{Prompt Construction}: Each agent constructs a prompt from statistics and metadata $M$, effective features from $\text{FeedMem}_i^{(r-1)}$, and distilled guidance from $\text{ConMem}_i^{(r-1)}$ and $\text{GlobalMem}^{(r-1)}$.
\item \textbf{Feature Generation and Evaluation}: Conditioned on the prompt, the agent uses $\pi_{\theta}$ to propose a transformation, instantiates it as $f_i^{(r)}$, evaluates the resulting features under $\mathcal{E}$, and stores the feedback in $\text{FeedMem}_i^{(r)}$.
\item \textbf{Memory Update}: To guide the next round, the agent updates $\text{ProcMem}_i^{(r)}$ and $\text{ConMem}_i^{(r)}$ with attempted operations, effective transformations, and newly identified patterns.

\end{itemize}

At the end of iteration $r$, the system selects top-performing features generated by the activated agents and integrates them into the dataset. Specifically, each active agent $A_i \in \mathcal{A}^{(r)}$ applies $\texttt{TopN-Features}$ to $T_i^{(r)}$ to retain the highest-ranked features under $\mathcal{E}$, and the selected features are aggregated to expand the global dataset. This iterative selection-and-aggregation procedure accumulates high-quality transformations and yields progressive improvements in model performance.


\section{Experiments}

\begin{table*}[t]
\centering
\small
\setlength{\tabcolsep}{5pt}
\caption{Performance (AUC) of all methods on 16 classification datasets. Best results are in bold, second-best are underlined. Results are averaged across three random train-test splits using the XGBoost classifier. ``N/A'' indicates that the running time exceeded 12 hours.}
\label{main:table}
\begin{tabular}{lcccccccc}

\toprule
\multirow{2}{*}{Datasets} & \multirow{2}{*}{Base} & \multicolumn{3}{c}{Traditional Methods} & \multicolumn{3}{c}{LLM-based Methods} & \multirow{2}{*} {MALMAS} \\
\cmidrule(lr){3-5} \cmidrule(lr){6-8}
 & & DFS & AutoFeat & OpenFE & CAAFE & OCTree & LLMFE & \\
\midrule

Adult & 0.849{\scriptsize $\pm$0.009} & 0.857{\scriptsize $\pm$0.001} & 0.849{\scriptsize $\pm$0.009} & 0.849{\scriptsize $\pm$0.009} & \underline{0.868{\scriptsize $\pm$0.005}} & 0.845{\scriptsize $\pm$0.011} & 0.853{\scriptsize $\pm$0.011} & \textbf{0.875{\scriptsize $\pm$0.010}} \\
Balance & 0.908{\scriptsize $\pm$0.009} & 0.989{\scriptsize $\pm$0.009} & 0.908{\scriptsize $\pm$0.009} & 0.908{\scriptsize $\pm$0.009} & \textbf{1.000{\scriptsize $\pm$0.000}} & 0.933{\scriptsize $\pm$0.031} & \underline{0.994{\scriptsize $\pm$0.008}} & \textbf{1.000{\scriptsize $\pm$0.000}} \\
Bank & 0.869{\scriptsize $\pm$0.007} & 0.891{\scriptsize $\pm$0.014} & 0.869{\scriptsize $\pm$0.007} & \textbf{0.904{\scriptsize $\pm$0.003}} & 0.874{\scriptsize $\pm$0.022} & 0.873{\scriptsize $\pm$0.010} & 0.875{\scriptsize $\pm$0.012} & \underline{0.895{\scriptsize $\pm$0.002}} \\
Banknote & 0.995{\scriptsize $\pm$0.002} & \underline{0.998{\scriptsize $\pm$0.001}} & 0.994{\scriptsize $\pm$0.003} & \underline{0.998{\scriptsize $\pm$0.001}} & 0.993{\scriptsize $\pm$0.002} & 0.993{\scriptsize $\pm$0.004} & 0.991{\scriptsize $\pm$0.002} & \textbf{0.999{\scriptsize $\pm$0.001}} \\
Breast\_W & \underline{0.989{\scriptsize $\pm$0.002}} & 0.988{\scriptsize $\pm$0.003} & 0.989{\scriptsize $\pm$0.002} & 0.988{\scriptsize $\pm$0.003} & \textbf{0.992{\scriptsize $\pm$0.001}} & 0.986{\scriptsize $\pm$0.003} & \textbf{0.992{\scriptsize $\pm$0.002}} & \textbf{0.992{\scriptsize $\pm$0.001}} \\
Car\_Eval & 0.982{\scriptsize $\pm$0.004} & 0.978{\scriptsize $\pm$0.002} & 0.982{\scriptsize $\pm$0.004} & \underline{0.994{\scriptsize $\pm$0.004}} & 0.988{\scriptsize $\pm$0.006} & 0.975{\scriptsize $\pm$0.007} & 0.986{\scriptsize $\pm$0.004} & \textbf{0.999{\scriptsize $\pm$0.000}} \\
Cdc &0.863{\scriptsize $\pm$0.001} & 0.863{\scriptsize $\pm$0.002} & N/A & {0.864{\scriptsize $\pm$0.001}} & \underline{0.866{\scriptsize $\pm$0.002}} & 0.865{\scriptsize $\pm$0.007} & 0.864{\scriptsize $\pm$0.001} & \textbf{0.867{\scriptsize $\pm$0.001}}

\\
Credit\_G & 0.756{\scriptsize $\pm$0.004} & \underline{0.758{\scriptsize $\pm$0.011}} & 0.756{\scriptsize $\pm$0.004} & 0.755{\scriptsize $\pm$0.011} & 0.751{\scriptsize $\pm$0.008} & 0.754{\scriptsize $\pm$0.011} & 0.748{\scriptsize $\pm$0.018} & \textbf{0.775{\scriptsize $\pm$0.002}} \\
Heart & 0.915{\scriptsize $\pm$0.001} & 0.916{\scriptsize $\pm$0.008} & 0.914{\scriptsize $\pm$0.001} & \underline{0.920{\scriptsize $\pm$0.008}} & 0.909{\scriptsize $\pm$0.002} & 0.912{\scriptsize $\pm$0.007} & 0.911{\scriptsize $\pm$0.004} & \textbf{0.923{\scriptsize $\pm$0.001}} \\
Jungle & 0.970{\scriptsize $\pm$0.000} & 0.975{\scriptsize $\pm$0.000} & 0.970{\scriptsize $\pm$0.000} & 0.980{\scriptsize $\pm$0.000} & \underline{0.983{\scriptsize $\pm$0.005}} & 0.972{\scriptsize $\pm$0.003} & 0.981{\scriptsize $\pm$0.006} & \textbf{0.993{\scriptsize $\pm$0.000}} \\

Myocardial & 0.802{\scriptsize $\pm$0.003} & 0.800{\scriptsize $\pm$0.010} & N/A & 0.803{\scriptsize $\pm$0.002} & \underline{0.805{\scriptsize $\pm$0.003}} & 0.803{\scriptsize $\pm$0.003} & \underline{0.805{\scriptsize $\pm$0.002}} & \textbf{0.809{\scriptsize $\pm$0.003}}

\\
Pima & 0.809{\scriptsize $\pm$0.007} & 0.810{\scriptsize $\pm$0.003} & 0.805{\scriptsize $\pm$0.006} & \underline{0.815{\scriptsize $\pm$0.008}} & 0.810{\scriptsize $\pm$0.007} & 0.810{\scriptsize $\pm$0.005} & 0.810{\scriptsize $\pm$0.008} & \textbf{0.823{\scriptsize $\pm$0.003}} \\
Student & 0.978{\scriptsize $\pm$0.001} & 0.977{\scriptsize $\pm$0.000} & 0.978{\scriptsize $\pm$0.001} & \underline{0.983{\scriptsize $\pm$0.000}} & 0.979{\scriptsize $\pm$0.001} & 0.978{\scriptsize $\pm$0.001} & 0.978{\scriptsize $\pm$0.001} & \textbf{0.984{\scriptsize $\pm$0.000}} \\
Churn & 0.829{\scriptsize $\pm$0.002} & \underline{0.833{\scriptsize $\pm$0.003}} & 0.829{\scriptsize $\pm$0.002} & 0.828{\scriptsize $\pm$0.001} & 0.827{\scriptsize $\pm$0.003} & 0.825{\scriptsize $\pm$0.002} & 0.829{\scriptsize $\pm$0.001} & \textbf{0.835{\scriptsize $\pm$0.001}} \\
Titanic & 0.843{\scriptsize $\pm$0.007} & 0.839{\scriptsize $\pm$0.005} & 0.843{\scriptsize $\pm$0.007} & 0.816{\scriptsize $\pm$0.005} & 0.843{\scriptsize $\pm$0.006} & 0.847{\scriptsize $\pm$0.004} & \underline{0.849{\scriptsize $\pm$0.004}} & \textbf{0.872{\scriptsize $\pm$0.008}} \\
Wine & 0.878{\scriptsize $\pm$0.001} & 0.885{\scriptsize $\pm$0.005} & 0.879{\scriptsize $\pm$0.002} & \textbf{0.891{\scriptsize $\pm$0.007}} & 0.878{\scriptsize $\pm$0.001} & 0.869{\scriptsize $\pm$0.006} & 0.879{\scriptsize $\pm$0.003} & \underline{0.886{\scriptsize $\pm$0.003}} \\
\midrule
\textbf{MeanRank} & 4.37 & 3.69 & 4.75 & \underline{3.12} & 3.57 & 4.81 & 3.75 & \textbf{1.12} \\

\bottomrule

\end{tabular}

\end{table*}

\begin{table*}[t]
\centering
\caption{Performance (NRMSE) of all methods on 7 regression datasets. Best results are in bold, second-best are underlined (Lower is better). Results are averaged across three random train-test splits using XGBoost regressor.}
\label{regression:table}
\begin{tabular}{lcccccc}
\toprule
\multirow{2}{*}{Datasets} & \multirow{2}{*}{Base} & \multicolumn{3}{c}{Traditional Methods} & \multicolumn{1}{c}{LLM-based Method} & \multirow{2}{*}{MALMAS} \\
\cmidrule(lr){3-5} \cmidrule(lr){6-6}
 & & DFS & AutoFeat & OpenFE & LLMFE & \\
\midrule
Airfoil & 0.015{\scriptsize $\pm$0.001} & 0.016{\scriptsize $\pm$0.001} & 0.015{\scriptsize $\pm$0.001} & \underline{0.014{\scriptsize $\pm$0.000}} & 0.015{\scriptsize $\pm$0.001} & \textbf{0.013{\scriptsize $\pm$0.000}} \\
Bike & 0.230{\scriptsize $\pm$0.001} & 0.225{\scriptsize $\pm$0.003} & 0.230{\scriptsize $\pm$0.001} & \textbf{0.213{\scriptsize $\pm$0.002}} & 0.225{\scriptsize $\pm$0.003} & \underline{0.215{\scriptsize $\pm$0.001}} \\
Crab & 0.220{\scriptsize $\pm$0.003} & {0.217{\scriptsize $\pm$0.002}} & 0.220{\scriptsize $\pm$0.003} & \underline{0.214{\scriptsize $\pm$0.002}} & 0.218{\scriptsize $\pm$0.002} & \textbf{0.213{\scriptsize $\pm$0.002}} \\
Insurance & 0.367{\scriptsize $\pm$0.006} & \underline{0.365{\scriptsize $\pm$0.010}} & 0.367{\scriptsize $\pm$0.006} & 0.381{\scriptsize $\pm$0.006} & 0.358{\scriptsize $\pm$0.007} & \textbf{0.355{\scriptsize $\pm$0.002}} \\

House & 0.173{\scriptsize $\pm$0.005} & 0.179{\scriptsize $\pm$0.019} & 0.173{\scriptsize $\pm$0.005} & \underline{0.160{\scriptsize $\pm$0.001}} & 0.165{\scriptsize $\pm$0.005} & \textbf{0.155{\scriptsize $\pm$0.004}} \\
Energy & 0.060{\scriptsize $\pm$0.002} & \textbf{0.046{\scriptsize $\pm$0.003}} & 0.060{\scriptsize $\pm$0.002} & 0.054{\scriptsize $\pm$0.003} & 0.058{\scriptsize $\pm$0.005} & \underline{0.050{\scriptsize $\pm$0.007}} \\

Medical &\underline{0.368{\scriptsize $\pm$0.003}} & 0.373{\scriptsize $\pm$0.002} & \underline{0.368{\scriptsize $\pm$0.003} }& 0.377{\scriptsize $\pm$0.001} & 0.370{\scriptsize $\pm$0.006} & \textbf{0.355{\scriptsize $\pm$0.003}} \\
\midrule

\textbf{MeanRank} & 3.86 & 3.29  & 3.86 & \underline{2.86} &3.14 & \textbf{1.29} \\
\bottomrule
\end{tabular}

\end{table*}

\subsection{Experimental Setup}
\subsubsection{Datasets}
Following prior work~\cite{baseline:1,baseline:6}, we evaluated our method on 16 classification and 7 regression datasets sourced from Kaggle and UCI. Following~\cite{auto:6}, we used a 6-4 train--test split and repeated each experiment three times with different seeds. 

\subsubsection{Baselines}
We compared MALMAS against a range of automated feature engineering baselines, including traditional methods such as AutoFeat~\cite{baseline:2}, OpenFE~\cite{baseline:4}, and DFS~\cite{baseline:3}, and LLM-based approaches such as CAAFE~\cite{baseline:1}, OCTree~\cite{auto:6}, and LLMFE~\cite{baseline:6}. The configurations of all baseline methods are detailed in Appendix~\ref{baseline:config}.

\subsubsection{Evaluation Metrics}

For classification tasks, we adopted the area under the AUC as the primary evaluation metric, and additionally reported accuracy (ACC) as a complementary measure, as shown in Appendix~\ref{acc:res}.
For regression tasks, we used the normalized root mean squared error (NRMSE) as the evaluation metric. Following prior work~\cite{baseline:6}, we also adopted mean rank as a global indicator to compare the overall effectiveness.

\subsubsection{MALMAS Configuration}
Across all experiments, MALMAS uses a fixed multi-agent configuration with $R{=}4$ iterative rounds, where agents generate candidate features, evaluate them. All methods are evaluated with the same downstream model, XGBoost~\cite{model:x}. LLM-based results in the main text use DeepSeekV3; additional details are deferred to Appendix~\ref{exper:info}.



\subsection{Overall Performance}

Table~\ref{main:table} summarizes the AUC performance of all evaluated methods across 16 classification datasets. Overall, MALMAS achieves the highest average AUC, consistently outperforming both traditional feature engineering methods and recent LLM-based approaches. MALMAS consistently improves upon the base model, ranking first or second on most benchmark datasets and exhibiting strong generalization across diverse real-world domains. Although LLM-based methods such as OCTree and LLMFE benefit from semantic-aware transformations, they still underperform compared to MALMAS in terms of overall average AUC. These results clearly and collectively underscore the effectiveness of memory-enhanced multi-agent collaboration in facilitating high-quality feature discovery.

As shown in Table~\ref{regression:table}, MALMAS also achieves the lowest mean NRMSE on almost regression tasks, indicating its strong and reliable feature generation capability beyond classification. Although some LLM-based baselines such as CAAFE and OCTree do not support regression, MALMAS still outperforms LLMFE by a large margin, confirming its advantage in continuous-value prediction.

\begin{figure}[t]
\centering
\includegraphics[width=0.85\columnwidth]{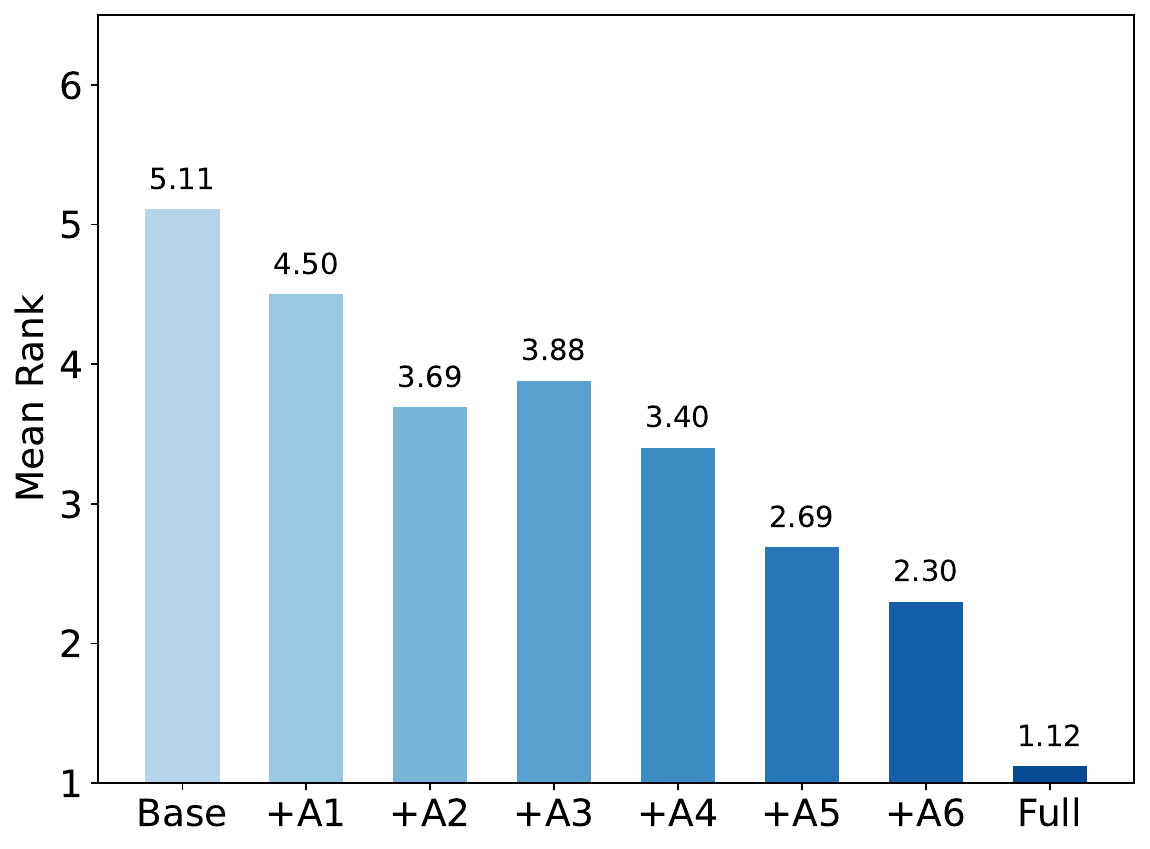} 
\caption{Mean rank (lower is better) across different ablation configurations of MALMAS.}
\vspace{-1.2mm}

\label{fig:3}
\end{figure}

\subsection{Ablation Study}
To gain deeper insights into the contributions of the multi-agent and memory modules, we further analyzed the results in Figure~\ref{fig:3}. From ``Base'' to ``+A6,'' the mean rank decreases from 5.11 to 2.30, indicating that expanding the agent pool broadens the feature search space and improves diversity, which in turn enhances downstream performance.

Beyond this, the ``Full'' configuration---which incorporates the memory module on top of all six agents---achieves a dramatic mean rank reduction to 1.12. This demonstrates the role of memory in accumulating cross-round information and refining feature-generation strategies. Specifically, procedural memory records attempted features to reduce redundancy, feedback memory stores downstream performance to guide the next round of exploration, and conceptual memory abstracts cross-round patterns summarized by the Summary Agent into a global conceptual memory shared across agents.

We observe a slight non-monotonicity from +A2 to +A3. This can plausibly occur because adding agents expands the candidate pool but may introduce higher-variance transformations that, under a fixed top-$N$ budget, occasionally replace more robust features; mean-rank aggregation is also sensitive to small dataset-level fluctuations.

\begin{figure}[t]
\centering
\includegraphics[width=0.85\columnwidth]{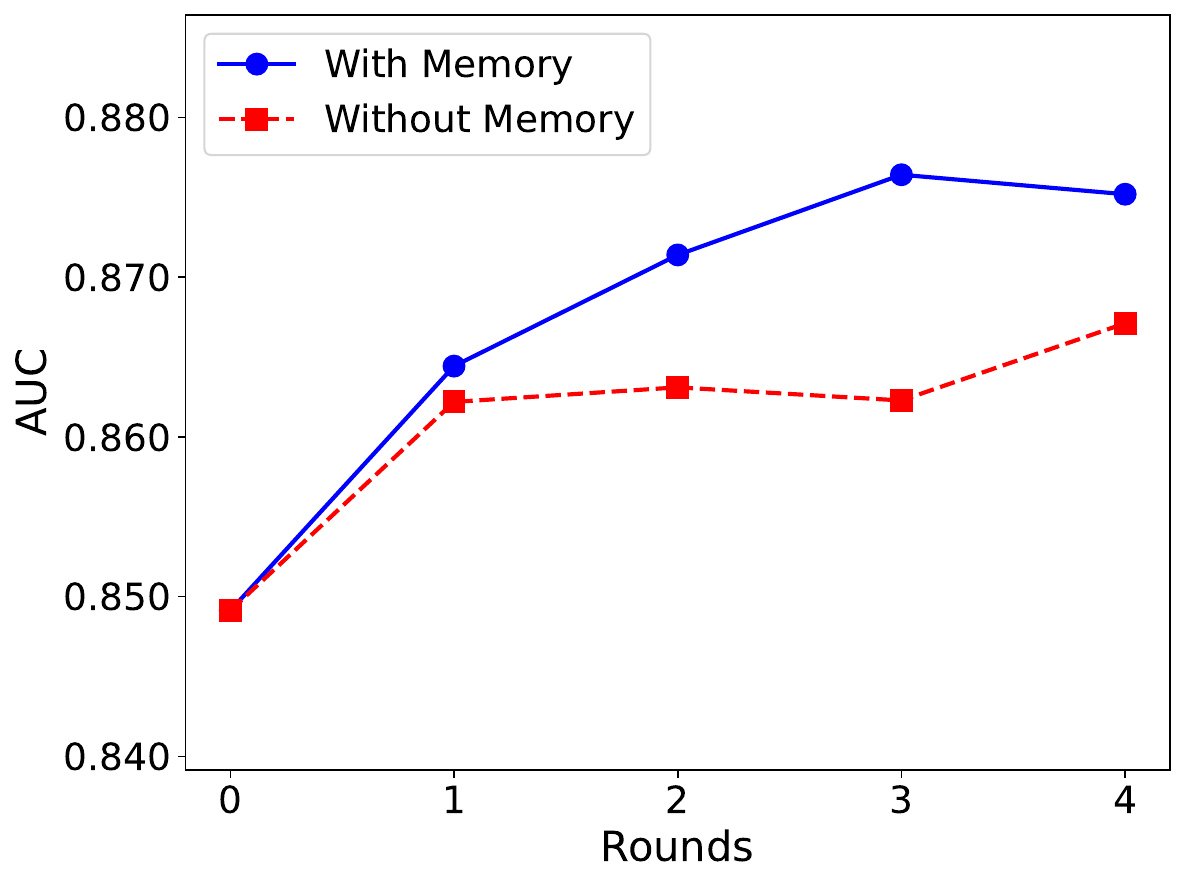} 
\caption{AUC performance on the Adult dataset across different rounds with and without memory.}

\vspace{-1.2mm}

\label{fig:4}
\end{figure}

\begin{table*}[t]
\centering
\caption{Classification performance (AUC) of H2O and DS-Agent on benchmark tabular datasets. 
``w/o'' indicates training with original features, while ``w/'' indicates training with our derived features. 
Results are reported as mean $\pm$ standard deviation over three runs.}

\label{end:table}

\begin{tabular}{lcccc}
\toprule
\multirow{2}{*}{Datasets} & \multicolumn{2}{c}{H2O} & \multicolumn{2}{c}{DS-Agent} \\
\cmidrule(lr){2-3} \cmidrule(lr){4-5}
 & w/o & w/ & w/o & w/ \\
\midrule
Adult                 & 0.876{\scriptsize $\pm$0.003} &\textbf{ 0.881{\scriptsize $\pm$0.001}} & {0.871}{\scriptsize $\pm$0.002} &\textbf{ 0.880{\scriptsize $\pm$0.001}} \\
Bank &                  0.864{\scriptsize $\pm$0.010} & \textbf{{0.899}{\scriptsize $\pm$0.029}} & 0.866{\scriptsize $\pm$0.012} & \textbf{0.892{\scriptsize $\pm$0.021} }\\
Breast\_W             & 0.989{\scriptsize $\pm$0.002} & \textbf{0.992{\scriptsize $\pm$0.003}} & {0.985}{\scriptsize $\pm$0.004} &\textbf{ 0.990{\scriptsize $\pm$0.002}} \\

Churn                   & 0.844{\scriptsize $\pm$0.003} & \textbf{0.846{\scriptsize $\pm$0.001}} & \textbf{{0.846}{\scriptsize $\pm$0.003}} & 0.845{\scriptsize $\pm$0.004} \\
Titanic                  & 0.859{\scriptsize $\pm$0.003} & \textbf{0.869{\scriptsize $\pm$0.001}} & {0.855}{\scriptsize $\pm$0.001} & \textbf{0.866{\scriptsize $\pm$0.003}} \\

\bottomrule
\end{tabular}
\vspace{-1mm}

\end{table*}

\subsection{Parameter Sensitivity}

The number of generation rounds $R$ is a key parameter in MALMAS, governing iterative feature refinement. Memory facilitates this process by guiding feature reuse, evaluation, and abstraction. Figure~\ref{fig:4} reports the AUC on the Adult dataset across rounds. With memory enabled, AUC increases from 0.85 to nearly 0.88 as $R$ grows, indicating that iterative generation can leverage accumulated feedback to uncover richer feature interactions. In contrast, without memory, performance plateaus after the first two rounds and slightly drops at round three, suggesting less directed exploration and limited gains in later rounds. Moreover, improvements diminish and plateau around rounds three to four, implying that MALMAS reaches a sufficiently rich feature set; dynamic scheduling could improve efficiency by using more agents/rounds early for exploration and focusing on high-value features later.

\subsection{Integration with Classical AutoML}\label{endtoend}

To further validate the effectiveness of our derived features in an end-to-end setting, we integrate them into classical AutoML pipelines based on H2O AutoML and DS-Agent~\cite{nollm:4,auto:2}. Table~\ref{end:table} reports the AUC performance of both methods on multiple tabular benchmark datasets. For each method, ``w/o'' uses the original features only, whereas ``w/'' augments them with our derived features.

In our experiments, H2O AutoML was run with a time budget of 2 hours for each run on each dataset, 
using five-fold cross-validation and ``AUC'' as the primary metric for model selection. 
For the DS-Agent pipeline, we used the same data splits and metric, 
and adopted DeepSeek-V3 as the LLM backbone to generate derived features.

Across all datasets, incorporating our derived features consistently improves the performance of both H2O AutoML and DS-Agent. 
This demonstrates that our feature derivation method integrates well with established end-to-end AutoML frameworks, yielding robust and reproducible gains.

\subsection{Discussion on Feature Generalization}
To assess whether MALMAS-generated features generalize beyond a single downstream model, we evaluated them across multiple classifiers, including XGBoost, LightGBM, Random Forest, and MLP~\cite{model:x,model:l,model:r}. Table~\ref{general:table} shows that MALMAS consistently achieves the highest AUC, with a mean rank of 1.00 (vs.\ 2.40 for the next-best method), indicating stable performance across diverse learning architectures and suggesting that the generated features capture broadly informative patterns rather than being tuned to a specific model. Compared with conventional approaches that may favor a particular algorithm, MALMAS remains effective under different decision boundaries and inductive biases; for example, both tree-based and neural models benefit from the enriched feature space. This cross-model consistency provides a reliable foundation in pipelines where model choice may vary across deployment settings. By mitigating dependence on any single learner, MALMAS can reduce repeated feature engineering and yield a more portable feature base for classification tasks.

\subsection{Computational and Token Cost}

To assess the feasibility of our method, we measured the average runtime and token usage of MALMAS on 16 classification datasets using the DeepSeek-V3 API as the LLM backbone. On average, each dataset required 0.452 hours of computation and 147.57k tokens for feature generation, with an estimated cost of \$0.17, as detailed in Appendix~\ref{Efficiency:detail}. These results indicate that MALMAS incurs modest overhead and can be readily embedded within existing AutoML pipelines.


\section{Conclusion}

We propose \textbf{MALMAS}, a memory-augmented multi-agent framework for automated feature generation, and validate its effectiveness through extensive experiments. By assigning distinct roles to specialized agents, MALMAS enables parallel and diverse exploration of the feature space, addressing the limitations of single-strategy approaches. Its memory module allows agents to retain useful signals and improve generation strategies across iterations. Together, these components provide a scalable and interpretable approach for producing high-quality, task-relevant features. We further present practical analyses demonstrating real-world applicability and efficiency.

\newpage
\section{Limitations}
MALMAS is designed for labeled tabular datasets and relies on downstream evaluation signals; its effectiveness may degrade when labels are scarce or evaluation budgets are limited. While our framework targets tabular feature engineering, its applicability to other modalities or structured domains remains unexplored. Moreover, as the candidate feature pool grows, repeated downstream training and validation can become a computational bottleneck, making performance sensitive to the available evaluation budget. Finally, although MALMAS provides transformation descriptions and memory traces, the overall LLM-driven generation process does not guarantee full interpretability of every derived feature.
\section{Ethical Considerations}
MALMAS is an automated feature generation framework for tabular data. Its outputs may inherit or amplify biases present in the input data, and the downstream evaluation signal may inadvertently favor transformations that correlate with sensitive attributes when such attributes are present or can be proxied. In addition, because MALMAS generates transformation programs, it may propose invalid or data-leaking features if the schema is ambiguous or the data pipeline is misconfigured. To mitigate these risks, we recommend applying strict schema constraints (e.g., explicitly marking protected attributes and leakage-prone fields), enforcing execution-time validation and leakage checks, and conducting fairness and privacy audits when deploying MALMAS in high-stakes settings. Finally, our experiments use publicly available datasets and do not involve human subjects.



\bibliography{custom}

\appendix


\clearpage

\section{Dataset Specifications}\label{datasets:info}

This section provides detailed specifications of all datasets used in our experiments, including the number of features, sample sizes, data sources, and official names. The datasets are categorized into classification and regression tasks as summarized in Table~\ref{tab:datasets}.

\begin{table*}[t]
\centering
\caption{Dataset Details and Sources}
\label{tab:datasets}
\begin{tabular}{llccc}
\toprule
\textbf{Datasets} & \textbf{Name} & \textbf{\#Features} & \textbf{\#Samples} & \textbf{Source} \\
\midrule
\multicolumn{5}{c}{\textbf{Classification Datasets}} \\
\midrule
Adult & Adult Census Income & 15 & 32561 & Kaggle \\
Balance & Balance Scale & 5 & 625 & Kaggle \\
Bank & Bank Marketing & 21 & 41188 & Kaggle \\
Banknote & Banknote Authentication & 5 & 1372 & UCI \\
Breast\_W & Breast Cancer Wisconsin (Original) & 9 & 699 & UCI \\
Car\_Eval & Car Evaluation & 7 & 1209 & Kaggle \\
Cdc     & diabetes health indicators dataset & 21 & 253680 &Kaggle\\
Credit\_G & German Credit Data & 21 & 1000 & Kaggle \\
Heart & Heart Disease & 12 & 918 & Kaggle \\
Jungle & Jungle Chess 2 Moves & 7 & 44819 & OpenML \\
Myocardial & myocardial infarction complications &111 &1700&UCI\\
Pima & Pima Indians Diabetes & 9 & 768 & Kaggle \\
Student & Student Performance Factors & 20 & 6607 & Kaggle \\
Churn & Telco Customer Churn & 21 & 7043 & Kaggle \\
Titanic & Titanic Dataset & 12 & 891 & Kaggle \\
Wine & Wine Quality & 13 & 6497 & Kaggle \\
\midrule
\multicolumn{5}{c}{\textbf{Regression Datasets}} \\
\midrule
Airfoile & Airfoil Self-Noise & 7 & 1504 & UCI \\
Bike & Bike Sharing & 13 & 17379 & UCI \\
Crab & Crab Age Prediction & 9 & 3893 & Kaggle \\
Insurance & Healthcare Insurance & 7 & 1338 & Kaggle \\
Housee & House Price Prediction & 36 & 1460 & Kaggle \\
Energy & Energy Efficiency & 9 & 768 & UCI \\
Medical & Medical Cost Personal Datasets & 7 & 1338 & Kaggle \\
\bottomrule
\end{tabular}

\end{table*}

To ensure a consistent and model-compatible feature representation, missing values in categorical features were imputed with a placeholder category ``NA'', while missing or infinite values in numerical features were replaced with zeros. All categorical variables were then encoded using \texttt{LabelEncoder}, mapping unseen categories during transformation to a fallback code of $-1$. This preprocessing ensured that all features were numerical and suitable for downstream learning algorithms.

\section{Implementation Details}
\subsection{Baseline Configurations}\label{baseline:config}
We implement and evaluate a variety of feature engineering baselines, spanning traditional symbolic approaches and recent LLM-based methods, to compare against our proposed MALMAS framework. All methods share the same downstream pipeline, using unified preprocessing and XGBoost as the default model to ensure fairness. Below, we summarize the baseline configurations.

\subsubsection*{AutoFeat}
AutoFeat is a symbolic feature engineering method that constructs new features using mathematical transformations such as polynomials, logarithms, and interactions. We adopt the open-source {autofeat} package and configure it to perform a single transformation step.

\subsubsection*{OpenFE}
OpenFE is an automated feature construction framework that combines feature boosting and pruning to identify informative transformations. We use the open-source {openfe} package with default settings.

\subsubsection*{Deep Feature Synthesis (DFS)}
DFS generates new features by applying aggregation and transformation operations over feature primitives. Following standard practice, we use {mean}, {standard deviation} as aggregators, and {add\_numeric}, {subtract\_numeric} as transformation primitives.

\subsubsection*{CAAFE}
CAAFE employs large language models to generate features via iterative sampling guided by control instructions. We use the official implementation with the number of iterations set to 10.

\subsubsection*{OCTree}
OCTree performs evolutionary search over operator trees to generate feature transformations. We use the official implementation for classification tasks and set the iteration count to 20.

\subsubsection*{LLMFE}
LLMFE uses a single-round prompt-based LLM generation pipeline without iterative feedback. We configure the method to sample 20 candidate features per seed.

All LLM-based methods are implemented using the DeepSeek API with a temperature of 1.0.

\subsection{Implementation Details of MALMAS}\label{exper:info}

MALMAS is a multi-agent, memory-augmented LLM framework for iterative feature generation. Unless otherwise specified, we use DeepSeekV3 as the backbone with temperature 1.0. We run $R{=}4$ interaction rounds, select the top-3 features per round, and apply a minimum effective-feature threshold of 2 during conceptual summarization. For all experiments, we use XGBoost as the downstream classifier with 500 trees and a learning rate of 0.02; for the relatively simple Car Evaluation and Banknote Authentication datasets, we reduce the number of trees to 50 to mitigate overfitting. This configuration is kept consistent across all baseline methods. We also evaluate GPT-4.1-mini on 16 classification datasets, with full results reported in Appendix~\ref{gpt41:detail}. All experiments are conducted on a machine with an Intel(R) Xeon(R) Gold 6326 CPU @ 2.90GHz and an NVIDIA RTX A6000. 






        




\section{Additional Results}\label{add:res}

\begin{table*}[t]
\centering
\small
\setlength{\tabcolsep}{5pt}
\caption{Performance (ACC) of all methods on 16 classification datasets. Best results are in bold, second-best are underlined. Results are averaged across three random train-test splits using the XGBoost classifier. ``N/A'' indicates that the running time exceeded 12 hours.}
\label{acc:table}
\begin{tabular}{lcccccccc}
\toprule
\multirow{2}{*}{Datasets} & \multirow{2}{*}{Base} & \multicolumn{3}{c}{Traditional Methods} & \multicolumn{3}{c}{LLM-based Methods} & \multirow{2}{*}{MALMAS} \\
\cmidrule(lr){3-5} \cmidrule(lr){6-8}
 & & DFS & AutoFeat & OpenFE & CAAFE & OCTree & LLMFE & \\
\midrule

Adult & 0.814{\scriptsize $\pm$0.015} & 0.810{\scriptsize $\pm$0.005} & 0.814{\scriptsize $\pm$0.015} & 0.814{\scriptsize $\pm$0.015} & \textbf{0.822{\scriptsize $\pm$0.007}} & 0.803{\scriptsize $\pm$0.014} & \underline{0.816{\scriptsize $\pm$0.013}} & \textbf{0.822{\scriptsize $\pm$0.013}} \\
Balance & 0.849{\scriptsize $\pm$0.008} & 0.961{\scriptsize $\pm$0.004} & 0.849{\scriptsize $\pm$0.008} & 0.849{\scriptsize $\pm$0.008} & \textbf{1.000{\scriptsize $\pm$0.000}} & 0.868{\scriptsize $\pm$0.020} & \underline{0.987{\scriptsize $\pm$0.019}} & \textbf{1.000{\scriptsize $\pm$0.000}} \\
Bank & 0.873{\scriptsize $\pm$0.009} & \underline{0.887{\scriptsize $\pm$0.008}} & 0.873{\scriptsize $\pm$0.009} & 0.884{\scriptsize $\pm$0.001} & 0.869{\scriptsize $\pm$0.010} & 0.881{\scriptsize $\pm$0.019} & 0.868{\scriptsize $\pm$0.009} & \textbf{0.897{\scriptsize $\pm$0.007}} \\
Banknote & 0.978{\scriptsize $\pm$0.003} & 0.978{\scriptsize $\pm$0.005} & 0.978{\scriptsize $\pm$0.001} & \underline{0.991{\scriptsize $\pm$0.004}} & 0.978{\scriptsize $\pm$0.000} & 0.973{\scriptsize $\pm$0.005} & 0.983{\scriptsize $\pm$0.003} & \textbf{0.993{\scriptsize $\pm$0.006}} \\
Breast\_W & 0.952{\scriptsize $\pm$0.009} & 0.954{\scriptsize $\pm$0.005} & 0.952{\scriptsize $\pm$0.009} & 0.956{\scriptsize $\pm$0.007} & \textbf{0.967{\scriptsize $\pm$0.004}} & 0.956{\scriptsize $\pm$0.002} & 0.958{\scriptsize $\pm$0.004} & \underline{0.960{\scriptsize $\pm$0.003}} \\
Car\_Eval & 0.901{\scriptsize $\pm$0.003} & 0.917{\scriptsize $\pm$0.006} & 0.901{\scriptsize $\pm$0.003} & \underline{0.961{\scriptsize $\pm$0.012}} & 0.924{\scriptsize $\pm$0.019} & 0.899{\scriptsize $\pm$0.003} & 0.916{\scriptsize $\pm$0.018} & \textbf{0.989{\scriptsize $\pm$0.003}} \\

Cdc &0.857{\scriptsize $\pm$0.002} & 0.857{\scriptsize $\pm$0.001} & N/A & \textbf{0.862{\scriptsize $\pm$0.001}} & {0.859{\scriptsize $\pm$0.001}} & 0.855{\scriptsize $\pm$0.002} & \underline{0.861{\scriptsize $\pm$0.001}} & \underline{0.861{\scriptsize $\pm$0.001}}
\\

Credit\_G & 0.744{\scriptsize $\pm$0.016} & 0.746{\scriptsize $\pm$0.012} & 0.744{\scriptsize $\pm$0.016} & 0.738{\scriptsize $\pm$0.005} & \textbf{0.757{\scriptsize $\pm$0.006}} & 0.741{\scriptsize $\pm$0.018} & \underline{0.748{\scriptsize $\pm$0.024}} & 0.747{\scriptsize $\pm$0.010} \\
Heart & 0.843{\scriptsize $\pm$0.017} & 0.841{\scriptsize $\pm$0.005} & 0.843{\scriptsize $\pm$0.010} & \underline{0.858{\scriptsize $\pm$0.003}} & 0.831{\scriptsize $\pm$0.013} & 0.852{\scriptsize $\pm$0.010} & 0.841{\scriptsize $\pm$0.003} & \textbf{0.864{\scriptsize $\pm$0.004} } \\
Jungle & 0.855{\scriptsize $\pm$0.000} & 0.868{\scriptsize $\pm$0.002} & 0.855{\scriptsize $\pm$0.000} & {0.882{\scriptsize $\pm$0.003}} & \underline{0.901{\scriptsize $\pm$0.022}} & 0.866{\scriptsize $\pm$0.017} & 0.895{\scriptsize $\pm$0.024} & \textbf{0.948{\scriptsize $\pm$0.001}} \\

Myocardial & 0.790{\scriptsize $\pm$0.006} & 0.791{\scriptsize $\pm$0.010} & N/A & 0.792{\scriptsize $\pm$0.002} & {0.794{\scriptsize $\pm$0.003}} & \underline{0.795{\scriptsize $\pm$0.003}} & \underline{0.795{\scriptsize $\pm$0.002}} & \textbf{0.799{\scriptsize $\pm$0.003}}
\\

Pima& 0.736{\scriptsize $\pm$0.011} & 0.753{\scriptsize $\pm$0.008} & 0.740{\scriptsize $\pm$0.005} & \underline{0.756{\scriptsize $\pm$0.007}} & 0.738{\scriptsize $\pm$0.013} & 0.754{\scriptsize $\pm$0.018} & 0.747{\scriptsize $\pm$0.011} & \textbf{0.759{\scriptsize $\pm$0.009}} \\
Student & 0.930{\scriptsize $\pm$0.003} & 0.929{\scriptsize $\pm$0.004} & 0.930{\scriptsize $\pm$0.003} & \underline{0.939{\scriptsize $\pm$0.002}} & 0.934{\scriptsize $\pm$0.002} & 0.930{\scriptsize $\pm$0.003} & 0.931{\scriptsize $\pm$0.003} & \textbf{0.943{\scriptsize $\pm$0.002}} \\
Churn & 0.787{\scriptsize $\pm$0.001} & \textbf{0.792{\scriptsize $\pm$0.002}} & 0.787{\scriptsize $\pm$0.001} & 0.787{\scriptsize $\pm$0.003} & 0.784{\scriptsize $\pm$0.003} & 0.789{\scriptsize $\pm$0.002} & 0.787{\scriptsize $\pm$0.002} & \underline{0.791{\scriptsize $\pm$0.004}} \\
Titanic & 0.768{\scriptsize $\pm$0.034} & \underline{0.782{\scriptsize $\pm$0.002}} & 0.768{\scriptsize $\pm$0.034} & 0.709{\scriptsize $\pm$0.051} & 0.774{\scriptsize $\pm$0.021} & 0.777{\scriptsize $\pm$0.036} & 0.781{\scriptsize $\pm$0.019} & \textbf{0.816{\scriptsize $\pm$0.007}} \\
Wine & 0.860{\scriptsize $\pm$0.001} & 0.868{\scriptsize $\pm$0.007} & 0.861{\scriptsize $\pm$0.002} & \textbf{0.872{\scriptsize $\pm$0.006}} & 0.864{\scriptsize $\pm$0.005} & 0.854{\scriptsize $\pm$0.005} & 0.863{\scriptsize $\pm$0.001} & \underline{0.869{\scriptsize $\pm$0.002}} \\
\midrule
MeanRank & 5.32 & 3.81 & 5.31 & \underline{3.31 }& 3.44 & 4.63 & 3.57 & \textbf{1.38} \\
\bottomrule

\end{tabular}

\end{table*}

\subsection{Supplementary Accuracy Results}\label{acc:res}

While AUC was used as the primary evaluation metric in our main experiments due to its robustness against class imbalance—common in many real-world classification datasets—we additionally report results based on Accuracy (ACC) to further validate the effectiveness and generalizability of the MALMAS framework.

Table~\ref{acc:table} reports the average ACC, with the following key observations:
\begin{itemize}
    \item MALMAS achieves the highest average accuracy, outperforming both traditional and LLM-based baselines.
    \item It ranks first on 11 out of 16 datasets and achieves top-2 performance on 4 datasets, demonstrating strong overall robustness.
    \item Compared to other LLM-based methods, MALMAS benefits from multi-agent collaboration and memory-guided prompt evolution, resulting in more diverse and relevant feature generation.
\end{itemize}

These results confirm that MALMAS delivers consistent classification performance not only under class-imbalance-aware metrics like AUC, but also under general-purpose metrics.

\subsection{Performance Analysis Using GPT-4.1-Mini}\label{gpt41:detail}
To validate the generalizability of our framework beyond a specific backbone, we also conducted experiments using GPT-4.1 Mini as the LLM for all LLM-based methods. Table~\ref{tab:gpt} reports the AUC performance across 16 classification datasets.

Overall, MALMAS achieves the best average rank (1.37), outperforming both traditional baselines (DFS, AutoFeat, OpenFE) and other LLM-based methods (CAAFE, OCTree, LLMFE). While the performance gap between different LLM-based methods narrows under a weaker backbone, MALMAS still maintains a consistent lead. This suggests that our multi-agent collaboration and memory mechanisms provide robust benefits even when the underlying LLM capacity is limited.

However, the overall performance of LLM-based methods, including MALMAS, tends to degrade slightly compared to results under stronger LLMs such as DeepSeekV3. For instance, methods like LLMFE and OCTree show more pronounced fluctuations and fall behind traditional methods on some datasets. This observation highlights an important insight: LLM-based feature generation is partially constrained by the expressive and reasoning capabilities of the underlying language model. Therefore, stronger LLMs contribute positively to semantic feature transformation, but architectural design remains critical for consistent gains.

These results serve as complementary evidence to our main experiments in the paper, demonstrating that MALMAS is not only effective with powerful LLMs, but also remains competitive and stable under smaller LLM configurations.

\begin{table*}[t]
\centering
\small
\setlength{\tabcolsep}{5pt}
\caption{Performance comparison ( AUC) of all methods across 16 classification datasets. Best results are in {bold}, second-best are {underlined}. Results are averaged across three random train-test splits. The GPT-4.1 Mini model was used as the backbone for the LLM-based methods. ``N/A'' indicates that the running time exceeded 12 hours.}

\label{tab:gpt}
\begin{tabular}{lcccccccc}
\toprule
\multirow{2}{*}{Datasets} & \multirow{2}{*}{Base} & \multicolumn{3}{c}{Traditional Methods} & \multicolumn{3}{c}{LLM-based Methods} & \multirow{2}{*}{MALMAS} \\
\cmidrule(lr){3-5} \cmidrule(lr){6-8}
 & & DFS & AutoFeat & OpenFE & CAAFE & OCTree & LLMFE & \\
\midrule
Adult & 0.849{\scriptsize $\pm$0.009} & 0.857{\scriptsize $\pm$0.001} & 0.849{\scriptsize $\pm$0.009} & 0.849{\scriptsize $\pm$0.009} & \textbf{0.868{\scriptsize $\pm$0.005}} & 0.845{\scriptsize $\pm$0.011} & 0.853{\scriptsize $\pm$0.011} & \underline{0.860{\scriptsize $\pm$0.013}} \\
Balance & 0.908{\scriptsize $\pm$0.009} & 0.989{\scriptsize $\pm$0.009} & 0.908{\scriptsize $\pm$0.009} & 0.908{\scriptsize $\pm$0.009} & \textbf{1.000{\scriptsize $\pm$0.000}} & 0.989{\scriptsize $\pm$0.007} & \underline{0.994{\scriptsize $\pm$0.008}} & \textbf{1.000{\scriptsize $\pm$0.000}} \\
Bank & 0.869{\scriptsize $\pm$0.007} & \underline{0.891{\scriptsize $\pm$0.014}}& 0.869{\scriptsize $\pm$0.007} & \textbf{0.904{\scriptsize $\pm$0.003}} & 0.873{\scriptsize $\pm$0.009} & 0.869{\scriptsize $\pm$0.007} & 0.867{\scriptsize $\pm$0.010} & {0.885{\scriptsize $\pm$0.011}} \\
Banknote & 0.995{\scriptsize $\pm$0.002} & \underline{0.998{\scriptsize $\pm$0.001}} & 0.994{\scriptsize $\pm$0.003} & \underline{0.998{\scriptsize $\pm$0.001}} & 0.993{\scriptsize $\pm$0.002} & 0.992{\scriptsize $\pm$0.004} & 0.992{\scriptsize $\pm$0.005} & \textbf{0.999{\scriptsize $\pm$0.001}} \\
Breast\_W & \underline{0.989{\scriptsize $\pm$0.002}} & 0.988{\scriptsize $\pm$0.003} & 0.989{\scriptsize $\pm$0.002} & 0.988{\scriptsize $\pm$0.003} & \textbf{0.991{\scriptsize $\pm$0.001}} & 0.989{\scriptsize $\pm$0.002} & \textbf{0.991{\scriptsize $\pm$0.002}} & \textbf{0.991{\scriptsize $\pm$0.001}} \\
Car & 0.982{\scriptsize $\pm$0.004} & 0.978{\scriptsize $\pm$0.002} & 0.982{\scriptsize $\pm$0.004} & \underline{0.994{\scriptsize $\pm$0.004}} & 0.993{\scriptsize $\pm$0.002} & 0.983{\scriptsize $\pm$0.007} & 0.991{\scriptsize $\pm$0.001} & \textbf{0.999{\scriptsize $\pm$0.000}} \\

Cdc &0.863{\scriptsize $\pm$0.001} & 0.863{\scriptsize $\pm$0.002} & N/A & {0.864{\scriptsize $\pm$0.001}} & \underline{0.865{\scriptsize $\pm$0.001}} & \textbf{0.866{\scriptsize $\pm$0.002}} & 0.864{\scriptsize $\pm$0.001} & \textbf{0.866{\scriptsize $\pm$0.001}}
\\
Credit\_G & 0.756{\scriptsize $\pm$0.004} & {0.758{\scriptsize $\pm$0.011}} & 0.756{\scriptsize $\pm$0.004} & 0.755{\scriptsize $\pm$0.011} & \underline{0.760{\scriptsize $\pm$0.009}} & 0.754{\scriptsize $\pm$0.011} & 0.756{\scriptsize $\pm$0.003} & \textbf{0.764{\scriptsize $\pm$0.004}} \\
Heart & 0.915{\scriptsize $\pm$0.001} & 0.916{\scriptsize $\pm$0.008} & 0.914{\scriptsize $\pm$0.001} & \underline{0.920{\scriptsize $\pm$0.008}} & 0.912{\scriptsize $\pm$0.004} & 0.912{\scriptsize $\pm$0.003} & 0.913{\scriptsize $\pm$0.004} & \textbf{0.923{\scriptsize $\pm$0.003}} \\
Jungle & 0.970{\scriptsize $\pm$0.000} & 0.975{\scriptsize $\pm$0.000} & 0.970{\scriptsize $\pm$0.000} & 0.980{\scriptsize $\pm$0.000} & \underline{0.983{\scriptsize $\pm$0.005}} & 0.974{\scriptsize $\pm$0.002} & 0.981{\scriptsize $\pm$0.006} & \textbf{0.988{\scriptsize $\pm$0.003}} \\

Myocardial & 0.802{\scriptsize $\pm$0.003} & 0.800{\scriptsize $\pm$0.010} & N/A & 0.803{\scriptsize $\pm$0.002} & {0.805{\scriptsize $\pm$0.002}} & 0.802{\scriptsize $\pm$0.003} & \underline{0.806{\scriptsize $\pm$0.001}} & \textbf{0.808{\scriptsize $\pm$0.003}}
\\
Pima & 0.809{\scriptsize $\pm$0.007} & 0.810{\scriptsize $\pm$0.003} & 0.805{\scriptsize $\pm$0.006} & \underline{0.815{\scriptsize $\pm$0.008}} & 0.801{\scriptsize $\pm$0.007} & 0.809{\scriptsize $\pm$0.007} & 0.813{\scriptsize $\pm$0.011} & \textbf{0.817{\scriptsize $\pm$0.007}} \\
Student & 0.978{\scriptsize $\pm$0.001} & 0.977{\scriptsize $\pm$0.000} & 0.978{\scriptsize $\pm$0.001} & \textbf{0.983{\scriptsize $\pm$0.000}} & 0.980{\scriptsize $\pm$0.001} & 0.978{\scriptsize $\pm$0.001} & 0.978{\scriptsize $\pm$0.004} & \underline{0.982{\scriptsize $\pm$0.001}} \\
Churn & 0.829{\scriptsize $\pm$0.002} & \underline{0.833{\scriptsize $\pm$0.003}} & 0.829{\scriptsize $\pm$0.002} & 0.828{\scriptsize $\pm$0.001} & 0.830{\scriptsize $\pm$0.002} & 0.827{\scriptsize $\pm$0.002} & 0.830{\scriptsize $\pm$0.003} & \textbf{0.834{\scriptsize $\pm$0.002}} \\
Titanic & 0.843{\scriptsize $\pm$0.007} & 0.839{\scriptsize $\pm$0.005} & 0.843{\scriptsize $\pm$0.007} & 0.816{\scriptsize $\pm$0.005} & 0.849{\scriptsize $\pm$0.011} & 0.843{\scriptsize $\pm$0.007} & 0.846{\scriptsize $\pm$0.010} & \textbf{0.855{\scriptsize $\pm$0.002}} \\
Wine & 0.878{\scriptsize $\pm$0.001} & \underline{0.885{\scriptsize $\pm$0.005}} & 0.879{\scriptsize $\pm$0.002} & \textbf{0.891{\scriptsize $\pm$0.007}} & 0.878{\scriptsize $\pm$0.005} & 0.878{\scriptsize $\pm$0.001} & 0.879{\scriptsize $\pm$0.001} & {0.882{\scriptsize $\pm$0.002}} \\
\midrule
\textbf{MeanRank} & 4.94 & 4.32 & 5.12 & 3.81 & \underline{3.44} & 5.12 & 3.87 & \textbf{1.37}
 \\
\bottomrule
\end{tabular}

\end{table*}

\begin{table*}[t] \centering
\small
\setlength{\tabcolsep}{5pt}
\caption{Performance comparison across different steps in the ablation study for each dataset using DeepSeekv3.}
\label{tab:steps}
\begin{tabular}{lcccccccc} \toprule 
Dataset & Base & +A1 & +A2 & +A3 & +A4 & +A5 &+A6&Full \\
\midrule Adult & 0.849{\scriptsize $\pm$0.009} & 0.853{\scriptsize $\pm$0.110} & 0.858{\scriptsize $\pm$0.009} & 0.867{\scriptsize $\pm$0.012} & 0.861{\scriptsize $\pm$0.015} & 0.865{\scriptsize $\pm$0.005} & 0.867{\scriptsize $\pm$0.020} & 0.875{\scriptsize $\pm$0.010} \\ Balance & 0.908{\scriptsize $\pm$0.009} & 0.908{\scriptsize $\pm$0.009} & 1.000{\scriptsize $\pm$0.000} & 1.000{\scriptsize $\pm$0.000} & 1.000{\scriptsize $\pm$0.000} & 1.000{\scriptsize $\pm$0.000} & 1.000{\scriptsize $\pm$0.000} & 1.000{\scriptsize $\pm$0.000} \\ Bank & 0.869{\scriptsize $\pm$0.007} & 0.878{\scriptsize $\pm$0.023} & 0.878{\scriptsize $\pm$0.006} & 0.878{\scriptsize $\pm$0.006} & 0.888{\scriptsize $\pm$0.011} & 0.881{\scriptsize $\pm$0.004} & 0.884{\scriptsize $\pm$0.011} & 0.895{\scriptsize $\pm$0.002} \\ Banknote & 0.995{\scriptsize $\pm$0.002} & 0.992{\scriptsize $\pm$0.003} & 0.995{\scriptsize $\pm$0.004} & 0.997{\scriptsize $\pm$0.003} & 1.000{\scriptsize $\pm$0.000} & 1.000{\scriptsize $\pm$0.000} & 1.000{\scriptsize $\pm$0.000} & 0.999{\scriptsize $\pm$0.001} \\ Breast\_W & 0.989{\scriptsize $\pm$0.002} & 0.989{\scriptsize $\pm$0.002} & 0.990{\scriptsize $\pm$0.001} & 0.989{\scriptsize $\pm$0.003} & 0.989{\scriptsize $\pm$0.003} & 0.990{\scriptsize $\pm$0.003} & 0.992{\scriptsize $\pm$0.003} & 0.992{\scriptsize $\pm$0.001} \\ Car\_Eval & 0.982{\scriptsize $\pm$0.004} & 0.991{\scriptsize $\pm$0.001} & 0.996{\scriptsize $\pm$0.001} & 0.998{\scriptsize $\pm$0.001} & 0.995{\scriptsize $\pm$0.002} & 0.997{\scriptsize $\pm$0.002} & 0.999{\scriptsize $\pm$0.001} & 0.999{\scriptsize $\pm$0.000} \\ 

Cdc & 0.863{\scriptsize $\pm$0.001} & 0.864{\scriptsize $\pm$0.001} & 0.865{\scriptsize $\pm$0.001} & 0.865{\scriptsize $\pm$0.002} & 0.865{\scriptsize $\pm$0.001} & 0.865{\scriptsize $\pm$0.001} & 0.866{\scriptsize $\pm$0.001} & 0.867{\scriptsize $\pm$0.001}

\\ Credit\_G & 0.756{\scriptsize $\pm$0.004} & 0.756{\scriptsize $\pm$0.001} & 0.752{\scriptsize $\pm$0.008} & 0.750{\scriptsize $\pm$0.013} & 0.757{\scriptsize $\pm$0.013} & 0.759{\scriptsize $\pm$0.009} & 0.765{\scriptsize $\pm$0.010} & 0.775{\scriptsize $\pm$0.002} \\ Heart & 0.915{\scriptsize $\pm$0.001} & 0.915{\scriptsize $\pm$0.001} & 0.910{\scriptsize $\pm$0.002} & 0.908{\scriptsize $\pm$0.010} & 0.911{\scriptsize $\pm$0.010} & 0.913{\scriptsize $\pm$0.008} & 0.907{\scriptsize $\pm$0.008} & 0.923{\scriptsize $\pm$0.001} \\ Jungle & 0.970{\scriptsize $\pm$0.000} & 0.972{\scriptsize $\pm$0.001} & 0.993{\scriptsize $\pm$0.003} & 0.978{\scriptsize $\pm$0.012} & 0.980{\scriptsize $\pm$0.010} & 0.987{\scriptsize $\pm$0.007} & 0.986{\scriptsize $\pm$0.005} & 0.993{\scriptsize $\pm$0.000} \\

Myocardial & 0.802{\scriptsize $\pm$0.003} & 0.802{\scriptsize $\pm$0.003} & 0.804{\scriptsize $\pm$0.001} & 0.804{\scriptsize $\pm$0.001} & 0.804{\scriptsize $\pm$0.002} & 0.806{\scriptsize $\pm$0.002} & 0.806{\scriptsize $\pm$0.001} & 0.809{\scriptsize $\pm$0.003}

\\ Pima & 0.809{\scriptsize $\pm$0.007} & 0.809{\scriptsize $\pm$0.011} & 0.811{\scriptsize $\pm$0.013} & 0.815{\scriptsize $\pm$0.009} & 0.817{\scriptsize $\pm$0.007} & 0.821{\scriptsize $\pm$0.005} & 0.824{\scriptsize $\pm$0.005} & 0.823{\scriptsize $\pm$0.003} \\ Student & 0.978{\scriptsize $\pm$0.001} & 0.979{\scriptsize $\pm$0.001} & 0.980{\scriptsize $\pm$0.001} & 0.980{\scriptsize $\pm$0.001} & 0.978{\scriptsize $\pm$0.001} & 0.980{\scriptsize $\pm$0.004} & 0.982{\scriptsize $\pm$0.000} & 0.984{\scriptsize $\pm$0.000} \\ Churn & 0.829{\scriptsize $\pm$0.002} & 0.833{\scriptsize $\pm$0.001} & 0.833{\scriptsize $\pm$0.001} & 0.831{\scriptsize $\pm$0.003} & 0.825{\scriptsize $\pm$0.009} & 0.833{\scriptsize $\pm$0.001} & 0.828{\scriptsize $\pm$0.007} & 0.835{\scriptsize $\pm$0.001} \\ Titanic & 0.843{\scriptsize $\pm$0.007} & 0.856{\scriptsize $\pm$0.016} & 0.859{\scriptsize $\pm$0.015} & 0.852{\scriptsize $\pm$0.006} & 0.865{\scriptsize $\pm$0.008} & 0.861{\scriptsize $\pm$0.007} & 0.868{\scriptsize $\pm$0.012} & 0.872{\scriptsize $\pm$0.008} \\ Wine & 0.878{\scriptsize $\pm$0.001} & 0.879{\scriptsize $\pm$0.001} & 0.881{\scriptsize $\pm$0.002} & 0.881{\scriptsize $\pm$0.001} & 0.882{\scriptsize $\pm$0.004} & 0.881{\scriptsize $\pm$0.004} & 0.883{\scriptsize $\pm$0.003} & 0.886{\scriptsize $\pm$0.003} \\ \midrule \textbf{MeanRank} & 5.11 & 4.50 & 3.69& {3.88} & 3.40 & 2.69 & 2.30 & {1.12} \\ \bottomrule \end{tabular} 

\end{table*}

\subsection{Details of the Ablation Study}\label{ablation:detail}

The ablation study results, as shown in Table~\ref{tab:steps}, provide a comprehensive evaluation of the incremental improvements in feature generation performance achieved by progressively adding agents and the memory module in the MALMAS. 

Overall Performance Improvement:
Starting from the ``Bas'' configuration, where the model is trained solely on raw features, the mean rank is 5.11. This serves as the baseline, highlighting the limitations of using untransformed raw features without any feature generation strategy. The introduction of each agent role (from +A1 to +A6) brings about noticeable improvements in performance, demonstrating the positive impact of specialized feature generation strategies. 

For example, when the first agent (+A1) is introduced, the mean rank improves to 4.50, but further improvements are not always linear. As additional agents are incorporated, the performance fluctuates, reaching a mean rank of 2.30 with the inclusion of all six agents (+A6). This trend suggests that the incorporation of agents with different roles progressively improves feature generation, as reflected by the decreasing mean rank at each step.

Impact of Memory Module:
The most significant improvement is observed when the full memory module is incorporated (Full configuration), resulting in the best performance with a mean rank of 1.12. The inclusion of the memory module, which integrates procedural memory, feedback memory, and conceptual memory, enables the system to iteratively refine feature generation strategies based on past experiences. This feedback loop allows the agents to adapt and enhance their strategies, contributing to the significant performance boost from +A6 to the Full configuration.

These results clearly demonstrate the effectiveness of the multi-agent and memory-augmented design of MALMAS. The gradual addition of agents and the final memory module significantly enhances the performance of model by enabling a more diverse and refined feature generation process. The final Full configuration, which integrates all components, shows the highest performance across all datasets, reaffirming the importance of both the multi-agent collaboration and the memory mechanism in driving high-quality feature discovery. The steady decrease in mean rank as each agent and memory component is added suggests that the MALMAS framework provides a robust and adaptive solution for feature generation.

\subsection{Routing Efficiency Analysis}\label{Routing}
To provide a more detailed analysis of the routing mechanism, we compare the proposed router with no-router and two controlled subset baselines (fixed-$K$ and random-$K$). We set $K=4$ for the subset baselines because the \textbf{average number of activated agents per round} in MALMAS is approximately 4, which makes the comparison fair. As shown in Table~\ref{tab:router-vs-baselines}, the router achieves comparable predictive performance to the no-router setting while requiring fewer tokens on average.

\begin{table*}[t]
\centering
\small
\setlength{\tabcolsep}{3pt}
\caption{Routing mechanism ablation: dataset-wise results on 16 classification datasets comparing no-router, fixed-$K$ and random-$K$ subset baselines (with $K=4$), and the proposed router.}
\label{tab:router-vs-baselines}
\begin{tabular}{lcccccccc}
\toprule
\textbf{dataset} & \textbf{no-router} & \makecell{\textbf{no-router}\\\textbf{tokens}} &
\makecell{\textbf{fixed-K}} & \makecell{\textbf{fixed-K}\\\textbf{tokens}} &
\makecell{\textbf{random-K}} & \makecell{\textbf{random-K}\\\textbf{tokens}} &
\textbf{router} & \makecell{\textbf{router}\\\textbf{tokens}} \\

\midrule
adult       & 0.876{\scriptsize $\pm$0.011} & 132 & 0.869{\scriptsize $\pm$0.010} & 104 & 0.860{\scriptsize $\pm$0.014} & 103 & 0.875{\scriptsize $\pm$0.010} & 111 \\
balance     & 1.000{\scriptsize $\pm$0.000} & 101 & 1.000{\scriptsize $\pm$0.000} & 79  & 1.000{\scriptsize $\pm$0.000} & 84  & 1.000{\scriptsize $\pm$0.000} & 88  \\
bank        & 0.895{\scriptsize $\pm$0.002} & 176 & 0.893{\scriptsize $\pm$0.004} & 149 & 0.890{\scriptsize $\pm$0.004} & 139 & 0.895{\scriptsize $\pm$0.002} & 141 \\
banknote    & 0.999{\scriptsize $\pm$0.001} & 113 & 0.999{\scriptsize $\pm$0.000} & 86  & 0.998{\scriptsize $\pm$0.001} & 82  & 0.999{\scriptsize $\pm$0.001} & 93  \\
breastw     & 0.992{\scriptsize $\pm$0.002} & 163 & 0.991{\scriptsize $\pm$0.002} & 153 & 0.989{\scriptsize $\pm$0.002} & 148 & 0.992{\scriptsize $\pm$0.001} & 141 \\
careval     & 0.999{\scriptsize $\pm$0.000} & 114 & 0.999{\scriptsize $\pm$0.000} & 95  & 0.998{\scriptsize $\pm$0.000} & 91  & 0.999{\scriptsize $\pm$0.000} & 97  \\
cdc         & 0.867{\scriptsize $\pm$0.001} & 184 & 0.866{\scriptsize $\pm$0.001} & 159 & 0.864{\scriptsize $\pm$0.001} & 150 & 0.867{\scriptsize $\pm$0.001} & 164 \\
credit      & 0.776{\scriptsize $\pm$0.001} & 218 & 0.770{\scriptsize $\pm$0.004} & 180 & 0.767{\scriptsize $\pm$0.004} & 173 & 0.775{\scriptsize $\pm$0.002} & 181 \\
heart       & 0.923{\scriptsize $\pm$0.001} & 121 & 0.920{\scriptsize $\pm$0.002} & 99  & 0.919{\scriptsize $\pm$0.001} & 100 & 0.923{\scriptsize $\pm$0.001} & 105 \\
jungle      & 0.991{\scriptsize $\pm$0.000} & 134 & 0.990{\scriptsize $\pm$0.001} & 110 & 0.986{\scriptsize $\pm$0.000} & 105 & 0.993{\scriptsize $\pm$0.000} & 118 \\
myocardial  & 0.809{\scriptsize $\pm$0.002} & 524 & 0.804{\scriptsize $\pm$0.002} & 489 & 0.804{\scriptsize $\pm$0.003} & 430 & 0.809{\scriptsize $\pm$0.003} & 463 \\
pima        & 0.824{\scriptsize $\pm$0.003} & 146 & 0.820{\scriptsize $\pm$0.003} & 130 & 0.817{\scriptsize $\pm$0.003} & 117 & 0.823{\scriptsize $\pm$0.003} & 126 \\
student     & 0.984{\scriptsize $\pm$0.000} & 184 & 0.982{\scriptsize $\pm$0.000} & 135 & 0.981{\scriptsize $\pm$0.000} & 130 & 0.984{\scriptsize $\pm$0.000} & 153 \\
chrn        & 0.836{\scriptsize $\pm$0.001} & 132 & 0.831{\scriptsize $\pm$0.004} & 90  & 0.830{\scriptsize $\pm$0.002} & 85  & 0.835{\scriptsize $\pm$0.001} & 94  \\
titanic     & 0.872{\scriptsize $\pm$0.007} & 173 & 0.870{\scriptsize $\pm$0.008} & 140 & 0.860{\scriptsize $\pm$0.010} & 138 & 0.872{\scriptsize $\pm$0.008} & 142 \\
wine        & 0.886{\scriptsize $\pm$0.003} & 165 & 0.884{\scriptsize $\pm$0.002} & 143 & 0.880{\scriptsize $\pm$0.004} & 140 & 0.886{\scriptsize $\pm$0.003} & 144 \\
\midrule
avg         & \textbf{0.908} & \textbf{173} & \textbf{0.906} & \textbf{146} & \textbf{0.903} & \textbf{138} & \textbf{0.908} & \textbf{148} \\
\bottomrule
\end{tabular}

\vspace{2mm}

\end{table*}

\subsection{Component-wise Memory Ablation}\label{Sec:MemAblation}
We further provide a component-wise ablation study of the memory module to answer which memory types are most crucial. Specifically, we compare \textit{No-Memory} with removing each memory component (i.e., \textit{-ProcMem}, \textit{-FeedMem}, \textit{-ConMem}, and \textit{-GlobalMem}) on the same 16 classification datasets. As summarized in Table~\ref{tab:memory-component-ablation}, enabling the full memory yields the best mean-rank, and removing any single component consistently degrades performance, with the largest drops observed for \textit{-GlobalMem} and \textit{-ConMem}.

\begin{table*}[t]
\centering
\setlength{\tabcolsep}{5pt}
\caption{Component-wise memory ablation on the same 16 classification datasets.}
\label{tab:memory-component-ablation}
\begin{tabular}{lcccccc}
\toprule
\textbf{Dataset} &
\textbf{No-Memory} &
\makecell{\textbf{-ProcMem}} &
\makecell{\textbf{-FeedMem}} &
\makecell{\textbf{-ConMem}} &
\makecell{\textbf{-GlobalMem}} &
\textbf{Full} \\
\midrule
adult       & 0.867{\scriptsize $\pm$0.020} & 0.872{\scriptsize $\pm$0.010} & 0.869{\scriptsize $\pm$0.011} & 0.871{\scriptsize $\pm$0.010} & 0.873{\scriptsize $\pm$0.011} & 0.875{\scriptsize $\pm$0.010} \\
balance     & 1.000{\scriptsize $\pm$0.000} & 1.000{\scriptsize $\pm$0.000} & 1.000{\scriptsize $\pm$0.000} & 1.000{\scriptsize $\pm$0.000} & 1.000{\scriptsize $\pm$0.000} & 1.000{\scriptsize $\pm$0.000} \\
bank        & 0.884{\scriptsize $\pm$0.011} & 0.892{\scriptsize $\pm$0.004} & 0.886{\scriptsize $\pm$0.003} & 0.890{\scriptsize $\pm$0.002} & 0.893{\scriptsize $\pm$0.001} & 0.895{\scriptsize $\pm$0.002} \\
banknote    & 1.000{\scriptsize $\pm$0.000} & 1.000{\scriptsize $\pm$0.000} & 1.000{\scriptsize $\pm$0.000} & 1.000{\scriptsize $\pm$0.000} & 1.000{\scriptsize $\pm$0.000} & 0.999{\scriptsize $\pm$0.001} \\
breastw     & 0.992{\scriptsize $\pm$0.003} & 0.992{\scriptsize $\pm$0.002} & 0.992{\scriptsize $\pm$0.001} & 0.992{\scriptsize $\pm$0.001} & 0.992{\scriptsize $\pm$0.001} & 0.992{\scriptsize $\pm$0.001} \\
careval     & 0.999{\scriptsize $\pm$0.001} & 0.999{\scriptsize $\pm$0.000} & 0.999{\scriptsize $\pm$0.000} & 0.999{\scriptsize $\pm$0.000} & 0.999{\scriptsize $\pm$0.000} & 0.999{\scriptsize $\pm$0.000} \\
cdc         & 0.866{\scriptsize $\pm$0.001} & 0.867{\scriptsize $\pm$0.002} & 0.866{\scriptsize $\pm$0.002} & 0.867{\scriptsize $\pm$0.001} & 0.867{\scriptsize $\pm$0.001} & 0.867{\scriptsize $\pm$0.001} \\
credit      & 0.765{\scriptsize $\pm$0.010} & 0.772{\scriptsize $\pm$0.004} & 0.769{\scriptsize $\pm$0.002} & 0.774{\scriptsize $\pm$0.001} & 0.773{\scriptsize $\pm$0.002} & 0.775{\scriptsize $\pm$0.002} \\
heart       & 0.907{\scriptsize $\pm$0.008} & 0.920{\scriptsize $\pm$0.003} & 0.914{\scriptsize $\pm$0.002} & 0.915{\scriptsize $\pm$0.001} & 0.920{\scriptsize $\pm$0.001} & 0.923{\scriptsize $\pm$0.001} \\
jungle      & 0.986{\scriptsize $\pm$0.005} & 0.990{\scriptsize $\pm$0.001} & 0.987{\scriptsize $\pm$0.001} & 0.991{\scriptsize $\pm$0.002} & 0.991{\scriptsize $\pm$0.001} & 0.993{\scriptsize $\pm$0.000} \\
myocardial  & 0.806{\scriptsize $\pm$0.001} & 0.807{\scriptsize $\pm$0.001} & 0.806{\scriptsize $\pm$0.002} & 0.807{\scriptsize $\pm$0.002} & 0.809{\scriptsize $\pm$0.003} & 0.809{\scriptsize $\pm$0.003} \\
pima         & 0.824{\scriptsize $\pm$0.005} & 0.824{\scriptsize $\pm$0.003} & 0.823{\scriptsize $\pm$0.004} & 0.824{\scriptsize $\pm$0.001} & 0.824{\scriptsize $\pm$0.001} & 0.823{\scriptsize $\pm$0.003} \\
student     & 0.982{\scriptsize $\pm$0.000} & 0.982{\scriptsize $\pm$0.000} & 0.982{\scriptsize $\pm$0.000} & 0.983{\scriptsize $\pm$0.000} & 0.984{\scriptsize $\pm$0.000} & 0.984{\scriptsize $\pm$0.000} \\
chrn        & 0.828{\scriptsize $\pm$0.007} & 0.833{\scriptsize $\pm$0.004} & 0.831{\scriptsize $\pm$0.003} & 0.832{\scriptsize $\pm$0.002} & 0.834{\scriptsize $\pm$0.002} & 0.835{\scriptsize $\pm$0.001} \\
titanic     & 0.868{\scriptsize $\pm$0.012} & 0.872{\scriptsize $\pm$0.009} & 0.870{\scriptsize $\pm$0.007} & 0.871{\scriptsize $\pm$0.006} & 0.871{\scriptsize $\pm$0.006} & 0.872{\scriptsize $\pm$0.008} \\
wine         & 0.883{\scriptsize $\pm$0.003} & 0.885{\scriptsize $\pm$0.002} & 0.884{\scriptsize $\pm$0.002} & 0.884{\scriptsize $\pm$0.001} & 0.885{\scriptsize $\pm$0.002} & 0.886{\scriptsize $\pm$0.003} \\
\midrule
Mean-rank & 3.5 & 2.0 & 3.1 & 2.2 & 1.66 & 1.1 \\
\bottomrule
\end{tabular}
\end{table*}


\subsection{Details of the Parameter Sensitivity}\label{rounds}

\begin{figure*}[t]
\centering
\includegraphics[width=1\textwidth]{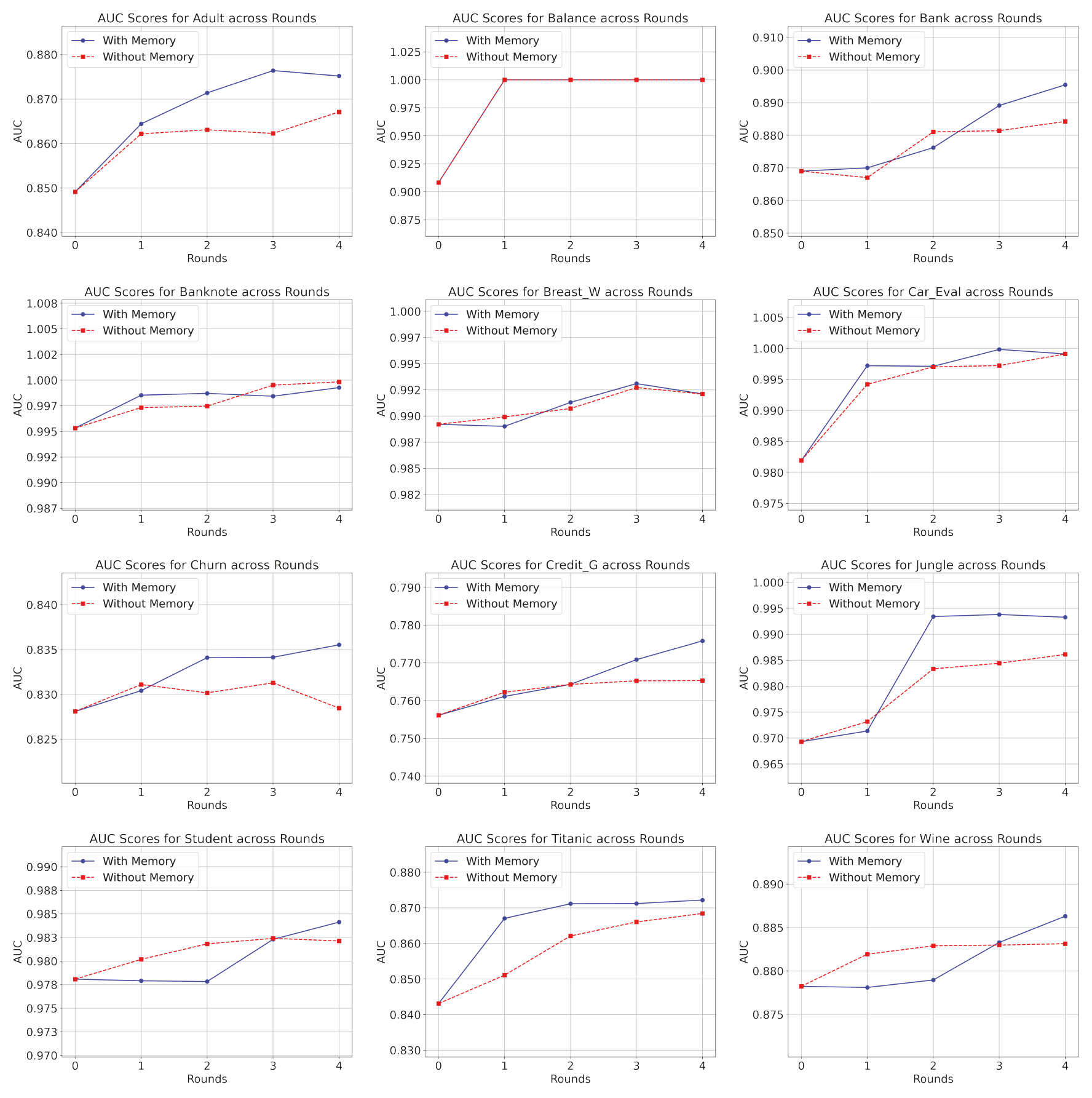} 
\caption{
AUC scores for multiple datasets across different rounds with and without memory 
}

\label{fig:5}
\end{figure*}

Figure~\ref{fig:5} presents the AUC scores across four feature generation rounds for multiple datasets, comparing models \textit{with} and \textit{without} memory. Each round corresponds to an additional iteration of feature generation and refinement by the MALMAS framework.
Key findings from this evaluation are summarized below:

\subsubsection*{Impact of Memory on Performance}
Models \textit{with memory} consistently outperform their \textit{without memory} counterparts across all rounds and most datasets. This demonstrates the effectiveness of memory-augmented feature generation in enabling progressive refinement. In contrast, models without memory show limited or stagnant improvement, particularly in early rounds, due to the lack of retained knowledge from prior iterations.

\subsubsection*{Performance Trends Across Rounds}
In datasets such as \textit{Adult}, \textit{Breast\_W}, and \textit{Jungle}, AUC scores steadily increase across rounds when memory is used, highlighting the cumulative benefits of iterative refinement. Without memory, the improvements are slower or negligible, reflecting the difficulty of building upon previously learned features without retention mechanisms.

\subsubsection*{Dataset-Specific Behavior}
On simpler datasets like \textit{Banknote}, both configurations achieve high AUC scores with a small performance gap. This suggests that memory may be less critical for less complex tasks. In contrast, for challenging datasets such as \textit{Churn} and \textit{Credit\_G}, the performance gap becomes more pronounced, indicating that memory is especially valuable for learning from intricate or nuanced data patterns.

\subsubsection*{Overall Implications}
The \textit{with memory} models exhibit a clear trend of continuous improvement across rounds, whereas \textit{without memory} models often plateau. This reflects the advantage of a memory-augmented framework in progressively enriching the feature space. The ability to incorporate and refine past knowledge plays a pivotal role in enhancing model performance, especially in complex scenarios.

In summary, these results provide strong empirical evidence for the benefits of memory in iterative feature generation. The memory-enhanced design of MALMAS significantly boosts learning capacity and generalization, enabling sustained performance gains over multiple rounds. This underscores the critical role of memory in advanced feature engineering systems.

\begin{table*}[t]
\centering
\setlength{\tabcolsep}{5pt}
\caption{AUC performance of all methods across four classifiers (Logistic Regression, XGBoost, LightGBM, Random Forest and MLP), averaged over 16 classification datasets. Best results are in bold; second-best are underlined.}

\label{general:table}

\begin{tabular}{lcccccccc}
\toprule
\multirow{2}{*}{Model} & \multirow{2}{*}{Base} & \multicolumn{3}{c}{Traditional Methods} & \multicolumn{3}{c}{LLM-based Methods} & \multirow{2}{*}{MALMAS} \\
\cmidrule(lr){3-5} \cmidrule(lr){6-8}
 & & DFS & AutoFeat & OpenFE & CAAFE & OCTree & LLMFE & \\
\midrule
Logistic Regression & 0.814 & 0.808 & 0.806 & \underline{0.823} & {0.819} & 0.815 & 0.814 & \textbf{0.845} \\
XGBoost & 0.890 & \underline{0.898} & 0.889 & 0.8903 & \underline{0.898} & 0.890 & 0.896 & \textbf{0.908} \\
LightGBM & 0.891 & {0.898} & 0.891 & 0.896 & \underline{0.899} & 0.891 & 0.898 & \textbf{0.906} \\

Random Forest & 0.887 & 0.892 & 0.889 & 0.889 & \underline{0.900} & 0.889 & \underline{0.900} & \textbf{0.905}  \\
MLP           & 0.863 & {0.867} & 0.862 & 0.866 & {0.869} & 0.861 &            \underline{0.870}         & \textbf{0.874}\\

\midrule
\textbf{MeanRank} & 5.20 & 3.60 & 5.80 & 3.80 & \underline{2.40} & 5.20 & {3.00} & \textbf{1.00} \\
\bottomrule
\end{tabular}

\end{table*}

\subsection{Generalization Experiment Setup}\label{general:detail}

For the generalization experiments across downstream models, all tree-based classifiers, including XGBoost, LightGBM, CatBoost, and Random Forest, were configured with 500 trees and a learning rate of 0.02, where applicable. Other hyperparameters were kept at their default settings as provided by each library. For two relatively simple datasets, car\_evaluation and banknote\_authentication, the number of trees was reduced to 50 to prevent overfitting due to their low data complexity. This configuration was consistently applied to all classification experiments to ensure fair evaluation of feature transferability across different model architectures.

Across all downstream classifiers—Logistic Regression (Table \ref{tab:lr}), LightGBM (Table~\ref{tab:l}), Random Forest (Table~\ref{tab:rf}) and MLP(Table~\ref{tab:mlp}) MALMAS consistently achieves the highest average AUC and the lowest mean rank. This indicates that the features generated by our multi-agent framework generalize robustly across diverse model architectures. Notably, while traditional methods such as OpenFE and AutoFeat perform competitively on simpler datasets, they fail to match MALMAS on complex ones. LLM-based baselines, including OCTree and LLMFE, benefit from semantic reasoning but still fall short in overall performance. These results confirm that the proposed framework is model-agnostic and maintains strong discriminative capability regardless of the downstream classifier.

\begin{table*}[t]
\centering
\small
\setlength{\tabcolsep}{5pt}
\caption{Performance (AUC) of all methods on 16 classification datasets. Best results are in bold, second-best are underlined. Results are averaged across three random train-test splits using the Logistic Regression classifier. ``N/A'' indicates that the running time exceeded 12 hours.}
\label{tab:lr}
\begin{tabular}{lcccccccc}
\toprule
\multirow{2}{*}{Datasets} & \multirow{2}{*}{Base} & \multicolumn{3}{c}{Traditional Methods} & \multicolumn{3}{c}{LLM-based Methods} & \multirow{2}{*}{MALMAS} \\
\cmidrule(lr){3-5} \cmidrule(lr){6-8}
 & & DFS & AutoFeat & OpenFE & CAAFE & OCTree & LLMFE & \\
\midrule
Adult & 0.777{\scriptsize $\pm$0.021} & 0.782{\scriptsize $\pm$0.020} & 0.801{\scriptsize $\pm$0.011} & 0.791{\scriptsize $\pm$0.022} & \underline{0.812{\scriptsize $\pm$0.011}} & 0.801{\scriptsize $\pm$0.003} &{0.778{\scriptsize $\pm$0.022} }& \textbf{0.813{\scriptsize $\pm$0.011}} \\
Balance & 0.863{\scriptsize $\pm$0.018} & {0.873{\scriptsize $\pm$0.028}} & 0.896{\scriptsize $\pm$0.018} & 0.867{\scriptsize $\pm$0.018} & \underline{0.999{\scriptsize $\pm$0.002}} & 0.877{\scriptsize $\pm$0.023} &{ 0.987{\scriptsize $\pm$0.016}} & \textbf{1.000{\scriptsize $\pm$0.000}} \\
Bank & 0.890{\scriptsize $\pm$0.016} &  0.896{\scriptsize $\pm$0.008} &0.896{\scriptsize $\pm$0.014} & 0.893{\scriptsize $\pm$0.010} & {0.886{\scriptsize $\pm$0.010}} &\underline{ 0.897{\scriptsize $\pm$0.002} }& 0.887{\scriptsize $\pm$0.008} & \textbf{0.901{\scriptsize $\pm$0.003}} \\
Banknote & {0.993{\scriptsize $\pm$0.001}} & {0.993{\scriptsize $\pm$0.001}} & \underline{0.998{\scriptsize $\pm$0.002}} & \textbf{0.999{\scriptsize $\pm$0.001}} & {0.995{\scriptsize $\pm$0.002}} & {0.990{\scriptsize $\pm$0.005}} & \underline{0.998{\scriptsize $\pm$0.002}} & \textbf{0.999{\scriptsize $\pm$0.001}} \\
Breast\_W & \underline{0.963{\scriptsize $\pm$0.002}} & \underline{0.963{\scriptsize $\pm$0.002}} &\underline{ 0.963{\scriptsize $\pm$0.002}} & 0.940{\scriptsize $\pm$0.012} & \underline{0.963{\scriptsize $\pm$0.002}} & 0.960{\scriptsize $\pm$0.004} & {0.962{\scriptsize $\pm$0.002}} & \textbf{0.971{\scriptsize $\pm$0.002}} \\
Car\_Eval & 0.689{\scriptsize $\pm$0.008} & 0.688{\scriptsize $\pm$0.005} & 0.689{\scriptsize $\pm$0.008} & \underline{0.797{\scriptsize $\pm$0.003}} & 0.716{\scriptsize $\pm$0.015} & 0.693{\scriptsize $\pm$0.014} & 0.697{\scriptsize $\pm$0.016} & \textbf{0.911{\scriptsize $\pm$0.054}} \\

Cdc &0.790{\scriptsize $\pm$0.002} & 0.794{\scriptsize $\pm$0.001} & N/A & {0.792{\scriptsize $\pm$0.001}} & {0.793{\scriptsize $\pm$0.001}} &  \underline{0.796{\scriptsize $\pm$0.001}} & \textbf{0.799{\scriptsize $\pm$0.001}} & \textbf{0.799{\scriptsize $\pm$0.001}}
\\
Credit\_G &{ 0.723{\scriptsize $\pm$0.009} }& 0.716{\scriptsize $\pm$0.005} & \underline{0.724{\scriptsize $\pm$0.013}} & 0.719{\scriptsize $\pm$0.006} & 0.722{\scriptsize $\pm$0.014} & 0.716{\scriptsize $\pm$0.005} & {0.712{\scriptsize $\pm$0.004}} & \textbf{0.729{\scriptsize $\pm$0.002}} \\
Heart & 0.849{\scriptsize $\pm$0.007} & 0.848{\scriptsize $\pm$0.006} & {0.708{\scriptsize $\pm$0.007}} & {0.851{\scriptsize $\pm$0.004}} & 0.831{\scriptsize $\pm$0.028} & {0.836{\scriptsize $\pm$0.001}} & \underline{0.853{\scriptsize $\pm$0.010}} & \textbf{0.857{\scriptsize $\pm$0.004}} \\
Jungle & 0.678{\scriptsize $\pm$0.003} & 0.678{\scriptsize $\pm$0.003} & 0.702{\scriptsize $\pm$0.002} &\underline{ 0.709{\scriptsize $\pm$0.013}} & {0.681{\scriptsize $\pm$0.003}} & 0.678{\scriptsize $\pm$0.003} & {0.685{\scriptsize $\pm$0.019}} & \textbf{0.710{\scriptsize $\pm$0.002}} \\

Myocardial & 0.764{\scriptsize $\pm$0.001} & 0.765{\scriptsize $\pm$0.010} & N/A &\underline{ 0.766{\scriptsize $\pm$0.002}} & \underline{{0.766{\scriptsize $\pm$0.002}} }& \textbf{0.767{\scriptsize $\pm$0.003}} & \underline{0.766{\scriptsize $\pm$0.001}} & \underline{0.766{\scriptsize $\pm$0.003}}
\\
Pima & \underline{0.779{\scriptsize $\pm$0.016}} & 0.719{\scriptsize $\pm$0.029} & {0.751{\scriptsize $\pm$0.037}} & {0.751{\scriptsize $\pm$0.015}} & 0.728{\scriptsize $\pm$0.074} & 0.765{\scriptsize $\pm$0.009} & 0.713{\scriptsize $\pm$0.054} & \textbf{0.781{\scriptsize $\pm$0.005}} \\
Student &{ 0.864{\scriptsize $\pm$0.012}} & 0.829{\scriptsize $\pm$0.004} & 0.861{\scriptsize $\pm$0.015} & \textbf{0.936{\scriptsize $\pm$0.013}} & 0.871{\scriptsize $\pm$0.027} & {0.864{\scriptsize $\pm$0.012}} & 0.852{\scriptsize $\pm$0.005} & \underline{0.879{\scriptsize $\pm$0.004}} \\
Churn & \underline{0.802{\scriptsize $\pm$0.003}} & 0.800{\scriptsize $\pm$0.004} & {0.795{\scriptsize $\pm$0.009}} & 0.799{\scriptsize $\pm$0.005} & 0.791{\scriptsize $\pm$0.005} &{ 0.799{\scriptsize $\pm$0.004}} & {0.796{\scriptsize $\pm$0.004}} & \textbf{0.806{\scriptsize $\pm$0.002}} \\
Titanic & \underline{0.785{\scriptsize $\pm$0.001}} & 0.782{\scriptsize $\pm$0.009} & 0.748{\scriptsize $\pm$0.005} & 0.741{\scriptsize $\pm$0.009} & {0.754{\scriptsize $\pm$0.039}} & \textbf{0.791{\scriptsize $\pm$0.010}} & {0.743{\scriptsize $\pm$0.005}} & \textbf{0.791{\scriptsize $\pm$0.022}} \\

Wine &\underline{ 0.816{\scriptsize $\pm$0.008} }& 0.810{\scriptsize $\pm$0.007} & {0.811{\scriptsize $\pm$0.004}} &{ 0.815{\scriptsize $\pm$0.005}} & {0.812{\scriptsize $\pm$0.006}} & 0.815{\scriptsize $\pm$0.006} & 0.812{\scriptsize $\pm$0.002} & \textbf{0.820{\scriptsize $\pm$0.006}} \\

\midrule
\textbf{Mean} & 0.814 & 0.808 & 0.806 & \underline{0.823} & {0.819} & 0.815 & 0.814 & \textbf{0.845} \\
\textbf{MeanRank} & 4.25 & 4.75 & 4.43 & \underline{3.68} & {4.37} & 3.81 &{4.31} & \textbf{1.12} \\

\bottomrule
\end{tabular}

\end{table*}
\begin{table*}[t] \centering
\small
\setlength{\tabcolsep}{5pt}
\caption{Performance (AUC) of all methods on 16 classification datasets. Best results are in bold, second-best are underlined. Results are averaged across three random train-test splits using the LightGBM classifier. ``N/A'' indicates that the running time exceeded 12 hours.}
\label{tab:l}

\begin{tabular}{lcccccccc} \toprule
\multirow{2}{*}{Datasets} & \multirow{2}{*}{Base} & \multicolumn{3}{c}{Traditional Methods} & \multicolumn{3}{c}{LLM-based Methods} & \multirow{2}{*}{MALMAS} \\
\cmidrule(lr){3-5} \cmidrule(lr){6-8}
 & & DFS & AutoFeat & OpenFE & CAAFE & OCTree & LLMFE & \\
\midrule
Adult & 0.852{\scriptsize $\pm$0.012} & 0.853{\scriptsize $\pm$0.006} & 0.852{\scriptsize $\pm$0.012} & 0.852{\scriptsize $\pm$0.012} & \textbf{0.863{\scriptsize $\pm$0.013}} & 0.840{\scriptsize $\pm$0.006} & 0.853{\scriptsize $\pm$0.014} & \underline{0.854{\scriptsize $\pm$0.024}} \\
Balance & 0.917{\scriptsize $\pm$0.008} & 0.990{\scriptsize $\pm$0.006} & 0.918{\scriptsize $\pm$0.008} & 0.917{\scriptsize $\pm$0.008} & \textbf{1.000{\scriptsize $\pm$0.000}} & 0.937{\scriptsize $\pm$0.043} & \underline{0.993{\scriptsize $\pm$0.009}} & \textbf{1.000{\scriptsize $\pm$0.000}} \\
Bank & 0.878{\scriptsize $\pm$0.008} & 0.896{\scriptsize $\pm$0.013} & 0.879{\scriptsize $\pm$0.009} & \underline{0.900{\scriptsize $\pm$0.006}} & 0.882{\scriptsize $\pm$0.019} & 0.882{\scriptsize $\pm$0.007} & 0.870{\scriptsize $\pm$0.011} & \textbf{0.902{\scriptsize $\pm$0.007}} \\
Banknote & 0.995{\scriptsize $\pm$0.001} & \underline{0.998{\scriptsize $\pm$0.002}} & 0.996{\scriptsize $\pm$0.000} & \textbf{0.999{\scriptsize $\pm$0.001}} & 0.995{\scriptsize $\pm$0.001} & 0.995{\scriptsize $\pm$0.001} & 0.994{\scriptsize $\pm$0.002} & \textbf{0.999{\scriptsize $\pm$0.001}} \\
Breast\_W & 0.990{\scriptsize $\pm$0.002} & 0.989{\scriptsize $\pm$0.002} & 0.989{\scriptsize $\pm$0.002} & 0.990{\scriptsize $\pm$0.003} & \textbf{0.993{\scriptsize $\pm$0.001}} & 0.988{\scriptsize $\pm$0.004} & \textbf{0.993{\scriptsize $\pm$0.001}} & \underline{0.992{\scriptsize $\pm$0.001}} \\
Car\_Eval & 0.981{\scriptsize $\pm$0.003} & 0.989{\scriptsize $\pm$0.002} & 0.981{\scriptsize $\pm$0.003} & \underline{0.997{\scriptsize $\pm$0.001}} & 0.989{\scriptsize $\pm$0.004} & 0.980{\scriptsize $\pm$0.001} & 0.986{\scriptsize $\pm$0.004} & \textbf{0.999{\scriptsize $\pm$0.000}} \\

Cdc &0.858{\scriptsize $\pm$0.002} & 0.861{\scriptsize $\pm$0.002} & N/A & {0.860{\scriptsize $\pm$0.001}} & \underline{0.863{\scriptsize $\pm$0.001}} & {0.862{\scriptsize $\pm$0.001}} & \textbf{0.864{\scriptsize $\pm$0.001}} & \textbf{0.864{\scriptsize $\pm$0.001}}

\\
Credit\_G & 0.750{\scriptsize $\pm$0.006} & 0.749{\scriptsize $\pm$0.012} & 0.751{\scriptsize $\pm$0.007} & \underline{0.762{\scriptsize $\pm$0.019}} & 0.755{\scriptsize $\pm$0.005} & 0.747{\scriptsize $\pm$0.005} & 0.744{\scriptsize $\pm$0.004} & \textbf{0.767{\scriptsize $\pm$0.014}} \\
Heart & 0.917{\scriptsize $\pm$0.003} & 0.917{\scriptsize $\pm$0.008} & 0.917{\scriptsize $\pm$0.004} & \textbf{0.922{\scriptsize $\pm$0.008}} & 0.912{\scriptsize $\pm$0.003} & 0.919{\scriptsize $\pm$0.009} & 0.914{\scriptsize $\pm$0.002} & \underline{0.920{\scriptsize $\pm$0.003}} \\
Jungle & 0.974{\scriptsize $\pm$0.000} & 0.979{\scriptsize $\pm$0.000} & 0.974{\scriptsize $\pm$0.000} & 0.984{\scriptsize $\pm$0.000} & \underline{0.988{\scriptsize $\pm$0.005}} & 0.977{\scriptsize $\pm$0.005} & 0.987{\scriptsize $\pm$0.006} & \textbf{0.996{\scriptsize $\pm$0.000}} \\

Myocardial & 0.799{\scriptsize $\pm$0.001} & 0.795{\scriptsize $\pm$0.010} & N/A &{ 0.799{\scriptsize $\pm$0.002}} & {{0.803{\scriptsize $\pm$0.002}} }& {0.800{\scriptsize $\pm$0.003}} & \underline{0.804{\scriptsize $\pm$0.001}} & \textbf{0.805{\scriptsize $\pm$0.003}}
\\

Pima & 0.805{\scriptsize $\pm$0.001} & 0.813{\scriptsize $\pm$0.005} & 0.805{\scriptsize $\pm$0.003} & \underline{0.814{\scriptsize $\pm$0.008}} & 0.804{\scriptsize $\pm$0.009} & 0.808{\scriptsize $\pm$0.010} & 0.813{\scriptsize $\pm$0.001} & \textbf{0.818{\scriptsize $\pm$0.005}} \\
Student & 0.981{\scriptsize $\pm$0.001} & 0.979{\scriptsize $\pm$0.000} & 0.981{\scriptsize $\pm$0.001} & \underline{0.984{\scriptsize $\pm$0.000}} & 0.981{\scriptsize $\pm$0.001} & 0.981{\scriptsize $\pm$0.001} & 0.981{\scriptsize $\pm$0.000} & \textbf{0.986{\scriptsize $\pm$0.000}} \\
Churn & 0.828{\scriptsize $\pm$0.002} & 0.828{\scriptsize $\pm$0.001} &\underline{ 0.827{\scriptsize $\pm$0.001}} & \underline{0.827{\scriptsize $\pm$0.002}} & 0.824{\scriptsize $\pm$0.004} & 0.825{\scriptsize $\pm$0.002} & \underline{0.827{\scriptsize $\pm$0.000}} & \textbf{0.828{\scriptsize $\pm$0.002}} \\
Titanic & 0.842{\scriptsize $\pm$0.011} & 0.842{\scriptsize $\pm$0.005} & 0.842{\scriptsize $\pm$0.011} & 0.843{\scriptsize $\pm$0.009} & 0.848{\scriptsize $\pm$0.004} & 0.849{\scriptsize $\pm$0.001} & \underline{0.854{\scriptsize $\pm$0.006}} & \textbf{0.876{\scriptsize $\pm$0.012}} \\
Wine & 0.886{\scriptsize $\pm$0.000} & \underline{0.890{\scriptsize $\pm$0.005}} & 0.886{\scriptsize $\pm$0.000} & \textbf{0.892{\scriptsize $\pm$0.006}} & 0.885{\scriptsize $\pm$0.003} & 0.874{\scriptsize $\pm$0.009} & 0.885{\scriptsize $\pm$0.004} & {0.886{\scriptsize $\pm$0.003}} \\
\midrule
\textbf{Mean} & 0.891 & \underline{0.898} & 0.891 & 0.896 & \underline{0.899} & 0.891 & 0.898 & \textbf{0.906}\\
\textbf{MeanRank} & 4.56 & 3.69 & 4.63 & \underline{2.94} & 3.19 & 4.31 & 3.43 & \textbf{1.32} \\

\bottomrule
\end{tabular}

\end{table*}

\begin{table*}[t]
\centering
\small
\setlength{\tabcolsep}{5pt}
\caption{Performance (AUC) of all methods on 16 classification datasets. Best results are in bold, second-best are underlined. Results are averaged across three random train-test splits using the RandomForest classifier. ``N/A'' indicates that the running time exceeded 12 hours.}
\label{tab:rf}
\begin{tabular}{lcccccccc}
\toprule
\multirow{2}{*}{Datasets} & \multirow{2}{*}{Base} & \multicolumn{3}{c}{Traditional Methods} & \multicolumn{3}{c}{LLM-based Methods} & \multirow{2}{*}{MALMAS} \\
\cmidrule(lr){3-5} \cmidrule(lr){6-8}
 & & DFS & AutoFeat & OpenFE & CAAFE & OCTree & LLMFE & \\
\midrule
Adult & 0.867{\scriptsize $\pm$0.007} & 0.851{\scriptsize $\pm$0.005} & 0.866{\scriptsize $\pm$0.005} & 0.866{\scriptsize $\pm$0.008} & \textbf{0.870{\scriptsize $\pm$0.003}} & 0.861{\scriptsize $\pm$0.005} &\underline{0.869{\scriptsize $\pm$0.010} }& \textbf{0.870{\scriptsize $\pm$0.011}} \\
Balance & 0.829{\scriptsize $\pm$0.007} & {0.972{\scriptsize $\pm$0.014}} & 0.843{\scriptsize $\pm$0.002} & 0.837{\scriptsize $\pm$0.007} & \textbf{1.000{\scriptsize $\pm$0.000}} & 0.871{\scriptsize $\pm$0.047} &\underline{ 0.982{\scriptsize $\pm$0.025}} & \textbf{1.000{\scriptsize $\pm$0.000}} \\
Bank & 0.891{\scriptsize $\pm$0.006} & 0.901{\scriptsize $\pm$0.011} & 0.891{\scriptsize $\pm$0.005} & \textbf{0.908{\scriptsize $\pm$0.007}} & 0.898{\scriptsize $\pm$0.005} & 0.885{\scriptsize $\pm$0.012} & 0.880{\scriptsize $\pm$0.007} & \underline{0.906{\scriptsize $\pm$0.007}} \\
Banknote & \textbf{1.000{\scriptsize $\pm$0.000}} & \textbf{1.000{\scriptsize $\pm$0.000}} & \textbf{1.000{\scriptsize $\pm$0.000}} & \textbf{1.000{\scriptsize $\pm$0.000}} & \textbf{1.000{\scriptsize $\pm$0.000}} & \textbf{1.000{\scriptsize $\pm$0.000}} & \textbf{1.000{\scriptsize $\pm$0.000}} & \textbf{1.000{\scriptsize $\pm$0.000}} \\
Breast\_W & 0.989{\scriptsize $\pm$0.003} & 0.988{\scriptsize $\pm$0.003} & 0.986{\scriptsize $\pm$0.003} & 0.989{\scriptsize $\pm$0.002} & \underline{0.990{\scriptsize $\pm$0.001}} & 0.987{\scriptsize $\pm$0.003} & \textbf{0.991{\scriptsize $\pm$0.003}} & \textbf{0.991{\scriptsize $\pm$0.000}} \\
Car\_Eval & 0.993{\scriptsize $\pm$0.001} & 0.989{\scriptsize $\pm$0.002} & 0.993{\scriptsize $\pm$0.001} & \underline{0.997{\scriptsize $\pm$0.001}} & 0.992{\scriptsize $\pm$0.004} & 0.990{\scriptsize $\pm$0.003} & 0.996{\scriptsize $\pm$0.001} & \textbf{0.999{\scriptsize $\pm$0.000}} \\

Cdc &0.832{\scriptsize $\pm$0.002} & 0.834{\scriptsize $\pm$0.002} & N/A & {0.834{\scriptsize $\pm$0.002}} & \underline{0.836{\scriptsize $\pm$0.001}} & {0.830{\scriptsize $\pm$0.001}} & \textbf{0.837{\scriptsize $\pm$0.001}} & \textbf{0.837{\scriptsize $\pm$0.001}}

\\
Credit\_G &\underline{ 0.768{\scriptsize $\pm$0.007} }& 0.759{\scriptsize $\pm$0.005} & 0.759{\scriptsize $\pm$0.008} & 0.767{\scriptsize $\pm$0.013} & 0.758{\scriptsize $\pm$0.019} & 0.755{\scriptsize $\pm$0.015} & {0.765{\scriptsize $\pm$0.005}} & \textbf{0.769{\scriptsize $\pm$0.004}} \\
Heart & 0.919{\scriptsize $\pm$0.006} & 0.916{\scriptsize $\pm$0.005} & \textbf{0.923{\scriptsize $\pm$0.006}} & \underline{0.921{\scriptsize $\pm$0.006}} & 0.917{\scriptsize $\pm$0.002} & \textbf{0.923{\scriptsize $\pm$0.007}} & 0.916{\scriptsize $\pm$0.002} & {0.919{\scriptsize $\pm$0.004}} \\
Jungle & 0.943{\scriptsize $\pm$0.000} & 0.955{\scriptsize $\pm$0.000} & 0.949{\scriptsize $\pm$0.000} & 0.963{\scriptsize $\pm$0.001} & \underline{0.979{\scriptsize $\pm$0.009}} & 0.954{\scriptsize $\pm$0.016} & {0.972{\scriptsize $\pm$0.010}} & \textbf{0.993{\scriptsize $\pm$0.000}} \\

Myocardial & 0.801{\scriptsize $\pm$0.001} & 0.799{\scriptsize $\pm$0.010} & N/A &{ 0.803{\scriptsize $\pm$0.002}} & {{0.803{\scriptsize $\pm$0.001}} }& {0.802{\scriptsize $\pm$0.002}} & \underline{0.804{\scriptsize $\pm$0.002}} & \textbf{0.805{\scriptsize $\pm$0.001}}
\\
Pima & 0.824{\scriptsize $\pm$0.008} & 0.823{\scriptsize $\pm$0.011} & \underline{0.827{\scriptsize $\pm$0.006}} & \textbf{0.829{\scriptsize $\pm$0.006}} & 0.825{\scriptsize $\pm$0.002} & 0.826{\scriptsize $\pm$0.008} & 0.826{\scriptsize $\pm$0.016} & \textbf{0.829{\scriptsize $\pm$0.005}} \\
Student &\underline{ 0.965{\scriptsize $\pm$0.001}} & 0.952{\scriptsize $\pm$0.001} & 0.964{\scriptsize $\pm$0.002} & 0.961{\scriptsize $\pm$0.001} & 0.962{\scriptsize $\pm$0.004} & \underline{0.965{\scriptsize $\pm$0.001}} & 0.961{\scriptsize $\pm$0.001} & \textbf{0.969{\scriptsize $\pm$0.000}} \\
Churn & 0.819{\scriptsize $\pm$0.001} & 0.810{\scriptsize $\pm$0.002} & \textbf{0.821{\scriptsize $\pm$0.001}} & 0.818{\scriptsize $\pm$0.000} & 0.810{\scriptsize $\pm$0.004} &\underline{ 0.819{\scriptsize $\pm$0.001}} & \textbf{0.821{\scriptsize $\pm$0.001}} & \textbf{0.821{\scriptsize $\pm$0.002}} \\
Titanic & 0.853{\scriptsize $\pm$0.011} & 0.831{\scriptsize $\pm$0.013} & 0.862{\scriptsize $\pm$0.004} & 0.821{\scriptsize $\pm$0.019} & {0.868{\scriptsize $\pm$0.019}} & 0.860{\scriptsize $\pm$0.006} & \underline{0.872{\scriptsize $\pm$0.014}} & \textbf{0.879{\scriptsize $\pm$0.022}} \\
Wine &\underline{ 0.899{\scriptsize $\pm$0.006} }& 0.897{\scriptsize $\pm$0.005} & \underline{0.899{\scriptsize $\pm$0.006}} & 0.893{\scriptsize $\pm$0.006} & \textbf{0.900{\scriptsize $\pm$0.006}} & 0.888{\scriptsize $\pm$0.013} & 0.896{\scriptsize $\pm$0.006} & 0.893{\scriptsize $\pm$0.006} \\

\midrule
\textbf{Mean} & 0.887 & 0.892 & 0.889 & 0.889 & \underline{0.900} & 0.889 & \underline {0.900} & \textbf{0.905} \\
\textbf{MeanRank} & 3.88 & 4.62 & 3.94 & 3.38 & \underline{2.94} & 4.25 &{2.87} & \textbf{1.44} \\

\bottomrule
\end{tabular}

\end{table*}

\begin{table*}[t]
\centering
\small
\setlength{\tabcolsep}{5pt}
\caption{Performance (AUC) of all methods on 16 classification datasets. Best results are in bold, second-best are underlined. Results are averaged across three random train-test splits using the MLP classifier. ``N/A'' indicates that the running time exceeded 12 hours.}
\label{tab:mlp}
\begin{tabular}{lcccccccc}
\toprule
\multirow{2}{*}{Datasets} & \multirow{2}{*}{Base} & \multicolumn{3}{c}{Traditional Methods} & \multicolumn{3}{c}{LLM-based Methods} & \multirow{2}{*}{MALMAS} \\
\cmidrule(lr){3-5} \cmidrule(lr){6-8}
 & & DFS & AutoFeat & OpenFE & CAAFE & OCTree & LLMFE & \\
\midrule
Adult & 0.834{\scriptsize $\pm$0.007} & 0.829{\scriptsize $\pm$0.007} & 0.836{\scriptsize $\pm$0.006} & 0.835{\scriptsize $\pm$0.008} & \underline{0.843{\scriptsize $\pm$0.007}} & 0.828{\scriptsize $\pm$0.006} & 0.833{\scriptsize $\pm$0.012} & \textbf{0.845{\scriptsize $\pm$0.005}} \\
Balance & 0.898{\scriptsize $\pm$0.008} & 0.952{\scriptsize $\pm$0.007} & 0.896{\scriptsize $\pm$0.006} & 0.897{\scriptsize $\pm$0.006} & \textbf{0.960{\scriptsize $\pm$0.000}} & 0.907{\scriptsize $\pm$0.024} & \underline{0.954{\scriptsize $\pm$0.009}} & \textbf{0.960{\scriptsize $\pm$0.000}} \\
Bank & 0.860{\scriptsize $\pm$0.011} & 0.869{\scriptsize $\pm$0.013} & 0.862{\scriptsize $\pm$0.010} & \textbf{0.876{\scriptsize $\pm$0.010}} & 0.862{\scriptsize $\pm$0.015} & 0.850{\scriptsize $\pm$0.008} & 0.861{\scriptsize $\pm$0.012} & \underline{0.875{\scriptsize $\pm$0.005}} \\
Banknote & {0.960{\scriptsize $\pm$0.000}} & {0.972{\scriptsize $\pm$0.000}} & {0.965{\scriptsize $\pm$0.000}} & \textbf{0.984{\scriptsize $\pm$0.000}} & \underline{0.980{\scriptsize $\pm$0.000}} & {0.967{\scriptsize $\pm$0.000}} & \underline{0.980{\scriptsize $\pm$0.000}} & \textbf{0.984{\scriptsize $\pm$0.000}} \\

Breast\_W & 0.950{\scriptsize $\pm$0.002} & 0.950{\scriptsize $\pm$0.002} & 0.950{\scriptsize $\pm$0.001} & 0.950{\scriptsize $\pm$0.001} & \textbf{0.953{\scriptsize $\pm$0.002}} & 0.948{\scriptsize $\pm$0.004} & \underline{0.952{\scriptsize $\pm$0.003}} & \underline{0.952{\scriptsize $\pm$0.002}} \\
Car\_Eval & 0.958{\scriptsize $\pm$0.000} & 0.954{\scriptsize $\pm$0.001} & 0.958{\scriptsize $\pm$0.000} & \underline{0.959{\scriptsize $\pm$0.000}} & 0.958{\scriptsize $\pm$0.000} & 0.956{\scriptsize $\pm$0.002} & 0.957{\scriptsize $\pm$0.000} & \textbf{0.960{\scriptsize $\pm$0.000}} \\

Cdc &0.818{\scriptsize $\pm$0.001} & 0.821{\scriptsize $\pm$0.002} & N/A & {0.820{\scriptsize $\pm$0.002}} & \underline{0.823{\scriptsize $\pm$0.001}} & {0.822{\scriptsize $\pm$0.001}} & \textbf{0.824{\scriptsize $\pm$0.001}} & \textbf{0.824{\scriptsize $\pm$0.001}}

\\
Credit\_G & 0.737{\scriptsize $\pm$0.009} & \underline{0.738{\scriptsize $\pm$0.007}} & 0.733{\scriptsize $\pm$0.005} & \underline{0.738{\scriptsize $\pm$0.009}} & 0.735{\scriptsize $\pm$0.005} & 0.728{\scriptsize $\pm$0.004} & 0.737{\scriptsize $\pm$0.003} & \textbf{0.743{\scriptsize $\pm$0.004}} \\
Heart & \textbf{0.890{\scriptsize $\pm$0.008}} & 0.886{\scriptsize $\pm$0.008} & \textbf{0.890{\scriptsize $\pm$0.008}} & \textbf{0.890{\scriptsize $\pm$0.008}} & 0.888{\scriptsize $\pm$0.009} & 0.887{\scriptsize $\pm$0.012} & 0.889{\scriptsize $\pm$0.009} & \underline{0.888{\scriptsize $\pm$0.007}} \\
Jungle & 0.934{\scriptsize $\pm$0.000} & 0.937{\scriptsize $\pm$0.001} & 0.934{\scriptsize $\pm$0.000} & 0.941{\scriptsize $\pm$0.001} & \underline{0.949{\scriptsize $\pm$0.007}} & 0.939{\scriptsize $\pm$0.007} & \underline{0.949{\scriptsize $\pm$0.006}} & \textbf{0.958{\scriptsize $\pm$0.000}} \\

Myocardial & 0.759{\scriptsize $\pm$0.004} & 0.755{\scriptsize $\pm$0.006} & N/A &{ 0.759{\scriptsize $\pm$0.002}} & {{0.763{\scriptsize $\pm$0.002}} }& {0.760{\scriptsize $\pm$0.004}} & \underline{0.764{\scriptsize $\pm$0.001}} & \textbf{0.766{\scriptsize $\pm$0.003}}
\\
Pima & 0.792{\scriptsize $\pm$0.006} & 0.790{\scriptsize $\pm$0.010} & 0.792{\scriptsize $\pm$0.004} & 0.792{\scriptsize $\pm$0.009} & 0.788{\scriptsize $\pm$0.004} & 0.792{\scriptsize $\pm$0.007} & \textbf{0.797{\scriptsize $\pm$0.010}} & \underline{0.793{\scriptsize $\pm$0.007}} \\
Student & 0.948{\scriptsize $\pm$0.001} & 0.947{\scriptsize $\pm$0.001} &\underline{0.949{\scriptsize $\pm$0.001}} & \textbf{0.950{\scriptsize $\pm$0.001}} &\underline{ 0.949{\scriptsize $\pm$0.000}} & 0.948{\scriptsize $\pm$0.001} & 0.948{\scriptsize $\pm$0.000} & \textbf{0.950{\scriptsize $\pm$0.001}} \\
Churn & \underline{0.796{\scriptsize $\pm$0.003}} & \textbf{0.797{\scriptsize $\pm$0.001}} & \textbf{0.797{\scriptsize $\pm$0.002}} & 0.794{\scriptsize $\pm$0.000} & 0.792{\scriptsize $\pm$0.001} & 0.795{\scriptsize $\pm$0.001} & \textbf{0.797{\scriptsize $\pm$0.001}} & \textbf{0.797{\scriptsize $\pm$0.002}} \\
Titanic & 0.820{\scriptsize $\pm$0.008} & 0.809{\scriptsize $\pm$0.004} & 0.818{\scriptsize $\pm$0.010} & 0.826{\scriptsize $\pm$0.009} & 0.822{\scriptsize $\pm$0.016} & 0.828{\scriptsize $\pm$0.008} & \underline{0.830{\scriptsize $\pm$0.010}} & \textbf{0.837{\scriptsize $\pm$0.016}} \\
Wine & 0.840{\scriptsize $\pm$0.004} & \underline{0.844{\scriptsize $\pm$0.005}} & 0.839{\scriptsize $\pm$0.004} & \textbf{0.847{\scriptsize $\pm$0.006}} & 0.838{\scriptsize $\pm$0.003} & 0.832{\scriptsize $\pm$0.008} & 0.840{\scriptsize $\pm$0.003} & 0.840{\scriptsize $\pm$0.005} \\

\midrule
\textbf{Mean} & 0.863 & {0.867} & 0.862 & 0.866 & {0.869} & 0.861 & \underline{0.870} & \textbf{0.874}\\
\textbf{MeanRank} & 3.87 & 3.75 & 4.00 & {2.75} & 2.68 & 4.26 &\underline{ 2.50} & \textbf{1.44} \\

\bottomrule
\end{tabular}

\end{table*}

\subsection{Additional Results on Time and Token Usage}\label{Efficiency:detail}
To evaluate the computational efficiency of our LLM-based feature generation process, we measured the time cost (in hours) and token usage (in thousands) on all 16 classification datasets. The experiments were conducted using the DeepSeek-V3 API as the LLM backbone and XGBoost as the downstream model, with all hyperparameters kept consistent with the main experiments. As shown in Table~\ref{tab:time_tokens}, the average generation time per dataset was approximately 0.452 hours, and the average token usage was around 147.57k tokens. These results demonstrate that our method is both time-efficient and computationally affordable, especially considering that feature generation is a one-time offline process. The variation in cost across datasets primarily reflects differences in metadata length and data complexity, but remains within acceptable bounds for real-world AutoML scenarios.

\begin{table*}[t]
\centering
\caption{Computation time and token usage on 16 classification datasets.}
\label{tab:time_tokens}
\begin{tabular}{lcccccccc}
\toprule
 & Adult & Balancee & Bank & Banknote & Breast\_W & Car\_Eval & Cdc & Credit\_G \\
\midrule
Time (h) & 0.22 & 0.12 & 0.21 & 0.11 & 0.20 & 0.11& 2.43& 0.31 \\
Tokens (k) & 111 & 88 & 141 & 93 & 141 & 97 & 164&181 \\

\bottomrule
\end{tabular}

\begin{tabular}{lccccccccc}
\\
\toprule
 & Heart & Jungle& Myocaridial & Pima & Student & Churn & Titanic & Wine & Mean \\
\midrule
Time (h) & 0.19 & 0.63& 1.32& 0.21 & 0.31 & 0.17 & 0.35 & 0.35 & 0.452 \\
Tokens (k) & 105 & 118 &463& 126 & 153 & 94 & 142 & 144 &147.57 \\
\bottomrule
\end{tabular}

\end{table*}



\end{document}